\documentclass[10pt,twocolumn,letterpaper]{article}

\usepackage{iccv}
\usepackage{times}
\usepackage{epsfig}
\usepackage{graphicx}
\usepackage{amsmath}
\usepackage{amssymb}
\usepackage{multirow}
\usepackage{lipsum}
\usepackage{adjustbox}
\usepackage{booktabs}
\usepackage{makecell}
\usepackage{tabu}
\usepackage{bm}
\usepackage{caption}
\usepackage{subcaption}

\usepackage{xcolor}
\usepackage{colortbl}
\usepackage{color}

\usepackage[pagebackref=true,breaklinks=true,letterpaper=true,colorlinks,bookmarks=false]{hyperref}

\usepackage{cleveref}
\iccvfinalcopy 

\ificcvfinal\fi

\definecolor{Gray}{gray}{0.95}
\definecolor{Grayy}{gray}{0.6}

\usepackage{soul}
\sethlcolor{Gray}
\newcommand{\PreserveBackslash}[1]{\let\temp=\\#1\let\\=\temp}
\newcolumntype{C}[1]{>{\PreserveBackslash\centering}p{#1}}
\newcolumntype{L}[1]{>{\PreserveBackslash\raggedright}p{#1}}
\ificcvfinal\fi

\usepackage{pifont}
\newcommand{\cmark}{\ding{51}}%
\newcommand{\xmark}{\ding{55}}%

\begin{document}


\title{T2M-HiFiGPT: Generating High Quality Human Motion from Textual Descriptions with Residual Discrete Representations}

\author{
\and
Second Author\\
Institution2\\
First line of institution2 address\\
{\tt\small secondauthor@i2.org}
}


\def\spaces{~~~~~~}
\author{\spaces{}Congyi Wang$^{\dagger}$}

\maketitle
\ificcvfinal\thispagestyle{empty}\fi

\begin{abstract}
In this study, we introduce T2M-HiFiGPT, a novel conditional generative framework for synthesizing human motion from textual descriptions. This framework is underpinned by a Residual Vector Quantized Variational AutoEncoder (RVQ-VAE) and a double-tier Generative Pretrained Transformer (GPT) architecture. We demonstrate that our CNN-based RVQ-VAE is capable of producing highly accurate 2D temporal-residual discrete motion representations. Our proposed double-tier GPT structure comprises a temporal GPT and a residual GPT. The temporal GPT efficiently condenses information from previous frames and textual descriptions into a 1D context vector. This vector then serves as a context prompt for the residual GPT, which generates the final residual discrete indices. These indices are subsequently transformed back into motion data by the RVQ-VAE decoder. To mitigate the exposure bias issue, we employ straightforward code corruption techniques for RVQ and a conditional dropout strategy, resulting in enhanced synthesis performance. Remarkably, T2M-HiFiGPT not only simplifies the generative process but also surpasses existing methods in both performance and parameter efficacy, including the latest diffusion-based and GPT-based models. On the HumanML3D~\cite{guo2022generating} and KIT-ML~\cite{plappert2016kit} datasets, our framework achieves exceptional results across nearly all primary metrics. We further validate the efficacy of our framework through comprehensive ablation studies on the HumanML3D~\cite{guo2022generating} dataset, examining the contribution of each component. Our findings reveal that RVQ-VAE is more adept at capturing precise 3D human motion with comparable computational demand compared to its VQ-VAE counterparts. As a result, T2M-HiFiGPT enables the generation of human motion with significantly increased accuracy, outperforming recent state-of-the-art approaches such as T2M-GPT~\cite{zhang2023generating}, Att-T2M~\cite{zhong2023attt2m} and FineMoGen~\cite{zhang2023finemogen}.

\end{abstract}

\section{Introduction}
In recent years, the burgeoning multimedia application market has propelled cross-modal human motion synthesis to the forefront of research in computer vision and graphics. A particularly intriguing branch of this field is text-driven human motion synthesis, which seeks to generate lifelike human motions that correspond seamlessly with provided textual descriptions. This technology holds the potential for transformative applications across various sectors, including the gaming industry and film production. For instance, in the realm of film and television production, the conventional method for creating new actions involves motion capture technology, which comes with a hefty price tag. The ability to automatically generate accurate 3D actions from textual prompts offers a promising alternative. It can yield rich, high-fidelity action data, significantly curtailing the time and resources traditionally required for animation production, thereby enhancing overall efficiency and cost-effectiveness.

Action generation based on natural language is challenging, mainly because (1) human motion in the real physical world is high-dimensional data with irregular spatial structure, highly dynamic temporal characteristics and many fine details. Direct text-driven synthesis in high-dimensional posture space (such as~\cite{li2017auto,valle2021transflower,tevet2022human} etc.) is challenging, and slight changes to high-dimensional vectors can lead to visible defects in the generated motion. (2) Actions and text come from different modalities. The model needs to learn the complex mapping relationship from language space to action space across modalities. For this, many works propose using autoencoders~\cite{ahuja2019language2pose,ghosh2021synthesis,tevet2022motionclip,yan2018mt} and VAE~\cite{zhong2023attt2m,zhang2023generating,petrovich2021action,petrovich2022temos,chen2023executing} to learn the joint embedding of language and action. MotionClip~\cite{tevet2022motionclip} aligns the action space with the CLIP~\cite{radford2021learning} space. ACTOR~\cite{petrovich2021action} and TEMOES~\cite{petrovich2022temos} respectively proposed Transformer-based VAEs for action-to-action and text-to-action. These works show promising performance under simple descriptions and are limited to generating high-quality actions when the text description becomes long and complex. Guo et al.~\cite{guo2022generating} and TM2T~\cite{guo2022tm2t} aim to generate action sequences with more challenging text descriptions. However, both methods are not direct, involve three stages of text-to-action generation, and sometimes fail to generate high-quality actions consistent with the text. Recently, diffusion-based models~\cite{ho2020denoising} have shown impressive results in image generation~\cite{ramesh2021zero}, then introduced into action generation by MDM~\cite{tevet2022human}, MotionDiffuse~\cite{zhang2022motiondiffuse},ReMoDiffuse~\cite{zhang2023remodiffuse},FineMoGen~\cite{zhang2023finemogen} and MotionLDM~\cite{chen2023executing}, and dominate the text-to-action generation task. However, we find that the inference speed improvement of diffusion-based methods~\cite{tevet2022human,zhang2022motiondiffuse,chen2023executing} may not be as significant as compared to classical methods (such as VQVAE~\cite{van2017neural}). Recent T2M-GPT~\cite{zhang2023generating} and Att-T2M~\cite{zhong2023attt2m} work, both use the VQVAE~\cite{van2017neural} model as the motion discrete representation; however, we find that real motion capture data contain very rich details, and simply using a single layer of discrete representation is not enough to fully and compactly express these motion details. Simply increasing the number of codebooks will lead to more redundancy, and the motion representation is not compact enough, which will also increase the training difficulty of the generation model due to large codebook sizes and data scarcity. In this work, unlike T2M-GPT~\cite{zhang2023generating} and Att-T2M~\cite{zhong2023attt2m}, we are inspired by the latest progress in learning discrete representations for generation~\cite{ao2022rhythmic,dhariwal2020jukebox,dieleman2018challenge,esser2021taming,lucas2022posegpt,ramesh2021zero,van2017neural,williams2020hierarchical,zeghidour2021soundstream,defossez2022high,wang2023neural}, and studied a simple framework based on Residual Vector Quantization Variational Autoencoder (RVQ-VAE)~\cite{zeghidour2021soundstream} and Valle-like~\cite{wang2023neural,lee2022autoregressive} structure for the compact and highly accurate discrete motion representation and text-to-action generation task. Specifically, we present a two-stage approach for synthesizing actions from textual descriptions. Initially, we employ a standard 1D dilated convolutional network to transform the action sequence into multi-layer residual discrete code indices. Subsequently, drawing inspiration from the speech generation model Valle~\cite{wang2023neural}, we introduce an innovative dual-tier GPT model designed to generate motion residual code sequences from pre-trained CLIP~\cite{radford2021learning} text embeddings. Our experimental analysis reveals that the RVQ-VAE model can effectively decrease the downsampling factor of motion data (from 4 to 8), while simultaneously reducing the computational time required for inference. This results in superior reconstruction outcomes on the test set. Moreover, as the sequence length for modeling is halved, the inference duration for the second stage is also proportionately diminished. Despite its straightforward nature, our method is capable of producing high-fidelity motion sequences that are intricately aligned with complex textual descriptions. In practical evaluations, our approach outperformed the GPT-based methods T2M-GPT~\cite{zhang2023generating} and Att-T2M~\cite{zhong2023attt2m} as well as recent diffusion-based methods ReMoDiffuse~\cite{zhang2023remodiffuse} and FineMoGen~\cite{zhang2023finemogen} on the widely recognized datasets HumanML3D~\cite{guo2022generating} and KIT-ML~\cite{plappert2016kit}. Notably, on the extensive HumanML3D~\cite{guo2022generating} dataset, our method achieved the highest accuracy, even exceeding the real data in terms of the consistency between text and generated motion, as measured by R-Precision and matching score. Through comprehensive experimentation, we have delved into this domain, and we hope that the insights and conclusions drawn from our work can contribute to future advancements in the field.

In summary, our contributions include:
\begin{enumerate}
    \item[(1)] Our findings indicate that the RVQ-VAE model with shared codebook, despite having comparable computational and parameter requirements to the VQ-VAE, exhibits superior performance in reconstructing 3D human motion data, thereby establishing itself as a more effective method for the discrete representation of arbitrary 3D human motion data.
    \item[(2)] We introduce T2M-HiFiGPT, a double-tier GPT model, to leverage the 2D discrete representation codes derived from RVQ-VAE for the text-to-motion task. It can achieve SOTA performance with much fewer parameters and comparable computational costs.
    \item[(3)] We have developed a suite of effective training strategies for T2M-HiFiGPT, which have significantly enhanced the model's matching accuracy and synthesis quality. 
    \item[(4)] We have conducted a large number of ablation experiments on all components of our method.
\end{enumerate}
\section{Related Work} 
\noindent \textbf{VQ-VAE and RVQ-VAE.} Vector Quantized Variational Autoencoder (VQ-VAE) is a variant of the VAE ~\cite{kingma2013auto}, was first proposed in~\cite{van2017neural}. VQ-VAE consists of an autoencoder architecture that aims to learn reconstructions with discrete representations. Recently, VQ-VAE has shown promising performance in generative tasks across different modalities, including image synthesis~\cite{ao2022rhythmic,esser2021taming,ramesh2021zero}, speech gesture generation~\cite{li2020learning, li2021ai, siyao2022bailando}. This study focuses on speech and music generation~\cite{dhariwal2020jukebox, dieleman2018challenge, wang2023neural}, etc. The success of VQ-VAE may be attributed to its ability to learn discrete representations and decouple priors. The naive training of VQ-VAE suffers from the problem of codebook collapse, i.e., only a few codes are activated, which significantly limits the performance of reconstruction and generation. To alleviate this problem, some techniques can be used during training, including optimizing the codebook along certain loss-gradient steps~\cite{van2017neural}, using Exponential Moving Average (EMA) for codebook updates~\cite{williams2020hierarchical}, resetting inactive codes during training (code reset)~\cite{williams2020hierarchical}), etc. RVQ-VAE~\cite{zeghidour2021soundstream} presents a further extension of VQ-VAE, a model that has shown success in speech synthesis~\cite{zeghidour2021soundstream, wang2023neural,agostinelli2023musiclm,borsos2023audiolm} and image synthesis~\cite{lee2022autoregressive}. RVQ-VAE introduces the concept of residual encoding. Any latent vector can accurately represent high-frequency signals, such as speech, by adding multiple VQ layers to infinitely approximate the latent vector. In this paper, we aim to utilize RVQ-VAE with \textit{shared} cookbooks to model motion signals for the first time. We discovered that by incorporating geometric, velocity, and acceleration loss functions during training, we can effectively reconstruct high-quality motion signals while reducing computational cost and representation length. This more compact novel representation not only makes it easier for us to train multimodal features related to text but also produces high-quality synthesized motion with fine details.
\vspace{1mm}

\noindent \textbf{Human body motion synthesis.} 
The field of human body motion synthesis has a rich research history~\cite{badler1993simulating}, with one of the most dynamic areas being the prediction of human body motion, which seeks to forecast future motion sequences based on past observed motions. Research efforts have primarily concentrated on the effective integration of spatial and temporal information, employing various models to generate deterministic future actions. These models include Recurrent Neural Networks (RNNs)~\cite{butepage2017deep,fragkiadaki2015recurrent,pavllo2018quaternet,martinez2017human}, Generative Adversarial Networks (GANs)~\cite{barsoum2018hp,hernandez2019human}, Graph Convolutional Networks (GCNs)~\cite{mao2019learning}, attention mechanisms~\cite{mao2020history}, and even simple Multi-Layer Perceptrons (MLPs)~\cite{bouazizi2022motionmixer,guo2023back}. Some approaches aim to produce a variety of actions using Variational Autoencoders (VAEs)~\cite{aliakbarian2020stochastic,komura2017recurrent,yan2018mt}, while others focus on generating "intermediate actions" that bridge the gap between past and future postures~\cite{duan2021single,harvey2018recurrent,harvey2020robust,kaufmann2020convolutional,tang2022real}. \cite{pavllo2018quaternet}. Notably, research has also explored the generation of action sequences from given trajectories, such as walking and running, and the synchronization of actions with music to create 3D dance movements~\cite{aristidou2022rhythm,chen2021choreomaster,lee2019dancing,li2020learning,li2021ai,siyao2022bailando}. For unconstrained generation,  \cite{yan2019convolutional} generates long sequences in a single step by transforming sequences of latent vectors sampled from Gaussian processes. In the realm of graphics, a significant body of work has been dedicated to kinematic control. Early efforts involved synthesizing high-quality human motion using motion graphs, which met user constraints through search and splicing techniques~\cite{barsoum2018hp,duan2021single}. Pioneering work by Holden et al.~\cite{holden2016deep} involved learning a convolutional autoencoder to reconstruct actions, with the resulting latent representation being utilized for the synthesis and editing of actions. Subsequent studies\cite{holden2017phase,starke2019neural} introduced a phase-function neural network for control tasks and a deep autoregressive framework for scene interaction behavior. More recent innovations~\cite{starke2022deepphase,li2022ganimator} include the reconstruction of actions through periodic features, which enhanced action synthesis, and the development of generative model methods for motion synthesis from a single sequence, inspired by SinGAN~\cite{shaham2019singan} to image synthesis.

\vspace{1mm}
\noindent \textbf{Text-based human action generation. } 
Text-based human action generation is a burgeoning field focused on creating 3D human actions from textual descriptions. Early work in this area includes Text2Action~\cite{ahn2018text2action}, which utilized an RNN-based model to produce actions conditioned on brief texts. Language2Pose~\cite{ahuja2019language2pose} took this further by employing a curriculum learning strategy to learn a joint embedding space of text and pose, enabling the generation of action sequences from text embeddings. Subsequent research by Ghosh et al.~\cite{ghosh2021synthesis} improved upon this by learning separate manifold representations for upper and lower body movements, yielding better results than Language2Pose~\cite{ahuja2019language2pose}. MotionCLIP~\cite{tevet2022motionclip} also sought to align text and action embeddings, advocating for the use of CLIP~\cite{radford2021learning} as a text encoder and incorporating rendered images for additional supervision. This approach showcased the potential to generate actions beyond known distributions and facilitated latent code editing, albeit with limitations in the quality and global translation of the generated action sequences. ACTOR~\cite{petrovich2021action} introduced a transformer-based VAE for generating actions from predefined classes in a non-autoregressive fashion. Building on ACTOR's architecture, TEMOS~\cite{petrovich2022temos} added a text encoder and produced more diverse action sequences, although it faced challenges with descriptions that fell outside its training distribution. TEACH~\cite{athanasiou2022teach} expanded upon TEMOS~\cite{petrovich2022temos} by generating sequences of actions from multiple natural language descriptions. The introduction of the large-scale dataset HumanML3D by Guo et al.~\cite{guo2022generating} marked a significant advancement, with the authors also recommending the incorporation of motion length prediction from text to generate motions of appropriate duration. TM2T~\cite{guo2022tm2t} tackled both text-to-motion and motion-to-text tasks, showing that joint training could yield additional improvements. With the rise of diffusion models, works like MDM~\cite{tevet2022human}, MotionDiffuse~\cite{zhang2022motiondiffuse}, ReMoDiffuse~\cite{zhang2023remodiffuse},FineMoGen~\cite{zhang2023finemogen} and MotionLDM~\cite{chen2023executing} have employed diffusion-based approaches for text-to-motion generation, achieving impressive results. However, these models often exhibit lower motion-text matching accuracy and slower inference speed due to the iterative nature of diffusion processes. Recent studies such as Att-T2M~\cite{zhong2023attt2m} and T2M-GPT~\cite{zhang2023generating} have both utilized the foundational VQ-VAE+GPT framework, with the former modeling the spatial relationships of limb parts using a Transformer structure~\cite{vaswani2017attention} within the VQ-VAE encoder, and the latter leveraging the classic GPT architecture to model the interplay between text and human motion. In contrast to their work, we have explored a more promising RVQ-VAE combined with a novel double-tier GPT structure that offers reduced computational costs and high-quality motion representation. With a straightforward design, this framework achieves superior performance through standard training methodologies, demonstrating its efficacy in the field of text-based human action generation.

\section{Method}
\label{sec:method}

\subsection{Method Overview}
\begin{figure*}[t]
\centering
\includegraphics[width=0.49\linewidth]{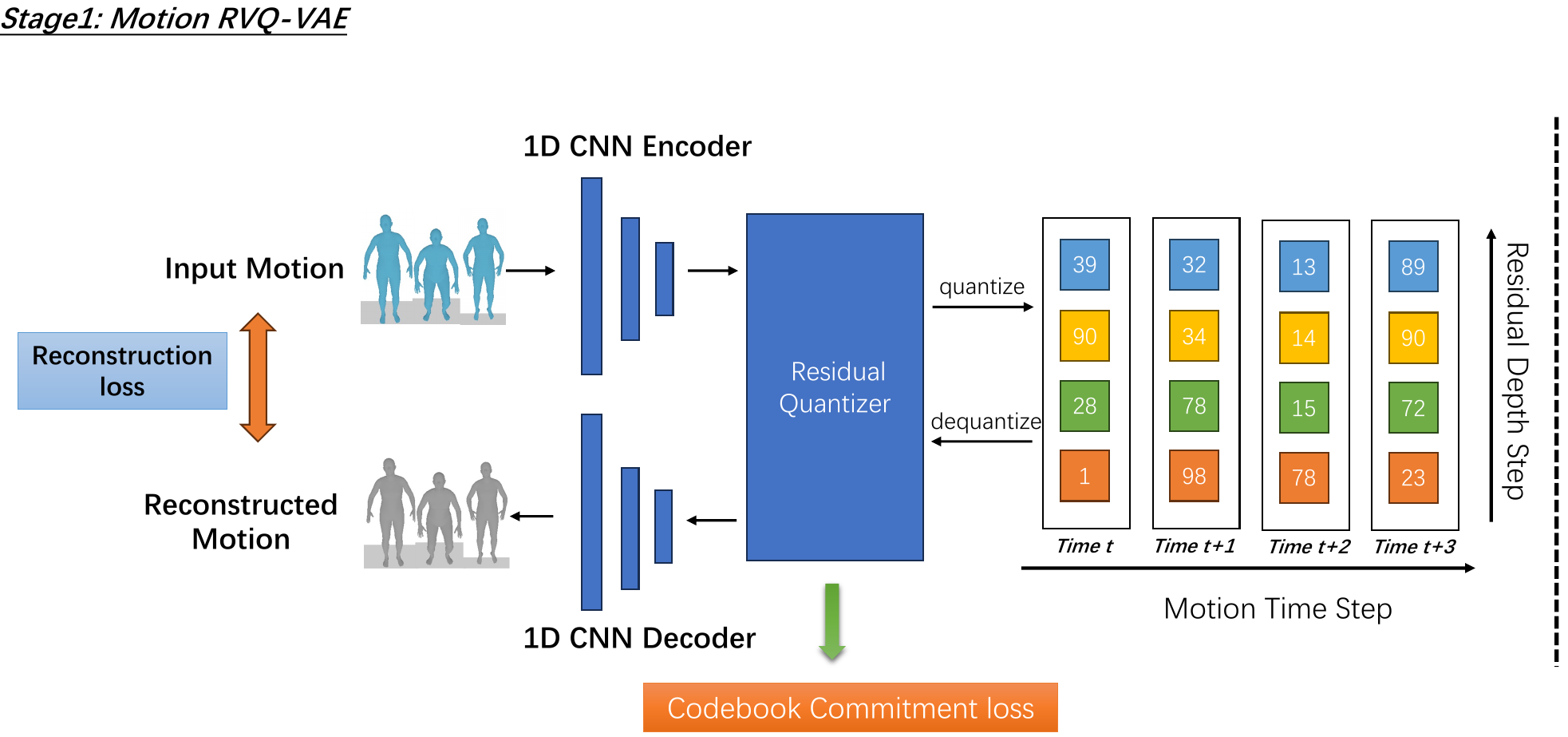}
\centering
\centering
\includegraphics[width=0.49\linewidth]{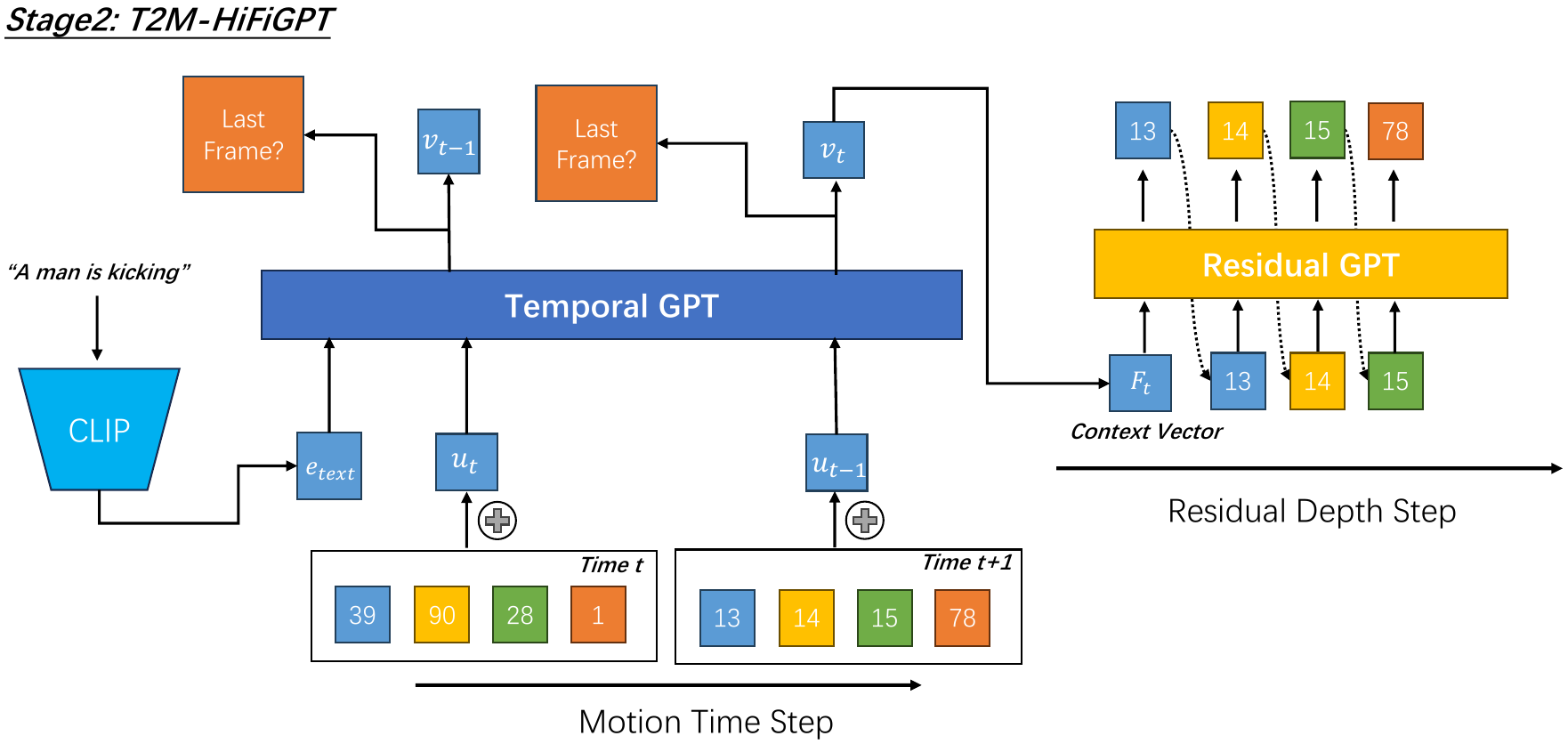}
\caption{The overall framework of T2M-HiFiGPT.}
\vspace{-10pt}
\label{fig:pipeline}
\end{figure*}

Our objective is to synthesize high-fidelity actions that are in harmony with the accompanying text descriptions. The overarching architecture of our framework is composed of two primary modules: Motion RVQ-VAE and T2M-HiFiGPT, as depicted in the ~\cref{fig:pipeline}. The initial module is tasked with learning the mapping between motion data and 2D residual discrete code sequences. Subsequently, the latter module is responsible for generating code indices derived from text descriptions. Utilizing the decoder within the Motion RVQ-VAE, we are able to reconstruct the original motion from these residual code indices. In Section 3.2, we delve into the intricacies of the Motion RVQ-VAE module. Following that, Section 3.3 is dedicated to the introduction of T2M-HiFiGPT. Finally, in Section 3.4, we discuss two pivotal training strategies designed to mitigate the issue of exposure bias and further enhance the accuracy of the synthesized actions.

\subsection{Motion RVQ-VAE}
RVQ-VAE was first proposed in SoundStream~\cite{zeghidour2021soundstream}, encodec~\cite{defossez2022high} and adopted for accurate reconstruction of speech signals, and later continued to be used for the text-to-image generation task~\cite{lee2022autoregressive}.  It can learn highly accurate discrete representations of the input signal by widening the residual depth, which can further be used in the second-stage generation models. In our scenario, given a motion sequence $X=[x_{1},x_{2},\ldots,x_{T}]$ with $x_{t} \in \mathcal{R}^{D}$, where $T$ is the number of frames and $D$ is the dimension of the motion data. We aim to recover the motion sequence through an autoencoder with 1D dilated convolution blocks similar to T2M-GPT~\cite{zhang2023generating} and a \textit{shared} learnable codebook containing $K$ codes $C=\{c_{k}|k=1,\ldots,K,c_{k} \in \mathcal{R}^{d}\}$, where $d$ is the dimension of codes. The overview of RVQ-VAE is presented in ~\cref{fig:pipeline}.

With encoder and decoder of the autoencoder denoted by $Enc$ and $Dec$, the latent feature $Z$ can be computed as $Z=Enc(X)$ with $Z = [z_{1}, z_{2},\ldots, z_{\lfloor T/l \rfloor}]$ and $z_{i} \in \mathcal{R}^{d}$, where $l$ represents the temporal downsampling rate of the encoder $Enc$, in our case, $l=8$. For latent feature $z_{i}$, the residual quantization through the \textit{shared} codebook $C$ is used to find the closest summation by greedily selecting $R$ codes in $C$, which can be properly formulated as: 
\begin{equation}
\label{eq:rvq_min}
    min_{c_{j} \in C,j=1,\ldots,R}\|z_{i}-\sum_{j=1}^{R}c_{j}\|
\end{equation}

where $R$ is the residual code depth. Note that increasing $R$ can lead to more accurate motion reconstruction, but at the cost of computational time and modeling complexity in the second stage. According to \cite{zeghidour2021soundstream}, \cref{eq:rvq_min} is solved by simple yet effective greedy search algorithms during the quantization. Specifically, each time we find the closest vector from codebook $C$ to match the current accumulated latent vector and repeat this process until reaching the coded depth $R$. Accordingly, the quantized latent feature $z_{i}^{\star}$ can be written as:
\begin{equation}
\label{eq:rvq_quant}
    z_{i}^{\star} = \sum_{j=1}^{R}c_{j}
\end{equation}
Note that when $R=1$, the formulation is exactly the vector quantization layer used in VQ-VAE~\cite{van2017neural}, As $R$ approaches infinity, $z_{i}^{\star}$ will infinitely approach the original latent vector $z_{i}$. Motion RVQ-VAE can produce a 2D discrete code matrix $S=\{s_{tw}|t=1,\ldots,\lfloor T/l \rfloor;w=1,\ldots,R;s_{tw} \in [0,\ldots,K-1]\}$ of shape $\lfloor T/l \rfloor \times R$ for the whole motion sequence $X$. The hierarchical decomposition of the motion signal shares the same concept with the spectrogram in speech processing, with low-level code indices representing coarse or low-frequency components and high-level code indices representing high-frequency fine details; accordingly, the residual code sequences possess strong \textbf{causal} property, which is very suited for GPT modeling.

\paragraph{Objective functions.} To optimize the above RVQ-VAE, we use the gradient descent with respect to the loss $L = L_{recon} + \beta L_{commit}$ with a multiplicative factor $\beta > 0$, the commitment loss $L_{commit}$ is defined as 
\begin{equation}
\label{eq:rvq_commit}
    L_{commit} = \sum_{j=1}^{R}\|Z-sg[c_{j}]\|_{2}^{2}
\end{equation}
where $sg[\cdot]$ is the stop-gradient operator, and the straight-through estimator~\cite{van2017neural} is used for the backpropagation through the RVQ quantization module. Note that $L_{commit}$ is the sum of quantization errors from every residual depth, not a single term in VQ-VAE. It aims to make $c_{j}$ sequentially decrease the quantization error of $Z$ as the depth increases. Thus, RVQ-VAE approximates the latent feature with a coarse-to-fine order.

For reconstruction, as suggested by~\cite{zhang2023generating}, we find that L1 smooth loss $LS$ performs best and an additional geometric regularization on the velocity, acceleration, and bone direction vectors further enhances the reconstruction quality. Let $X_{re}$ be the reconstructed motion of $X$, i.e., $X_{re}=Dec(Z)$, $V(X)$ be the velocity of $X$ where $V = [v_{1},v_{2},\ldots, v_{T-1}]$ with $v_{i} = x_{i+1}-x_{i}$, $A(X)$ be the acceleration of X where $A = [a_{2},a_{3},\ldots,a_{T-2}]$ with $a_{i} = x_{i+2}+x_{i}-2x_{i+1}$. Therefore, the objective of the reconstruction is as follows:
\begin{equation}
\label{eq:rvq_recon}
\begin{split}
    L_{recon} = LS(X,X_{re}) + \alpha_{1}LS(V(X),V(X_{re})) + \\ \alpha_{2}LS(A(X),A(X_{re})) + \\
    \alpha_{3}\sum_{(u,v) \in \text{joint-set}} \|(X(u)-X(v))-(X_{re}(u) - X_{re}(v))\|
\end{split}
\end{equation}

where $X(u)$ and $X(v)$ denote the specific 3D joint position, $X(u)-X(v)$ represents the bone direction vector from joint $v$ to joint $u$, $\alpha_{1,2,3}$ are hyper-parameters to balance all the losses above. To keep the training stable, the \textit{shared} codebook $C$ is updated by the exponential moving average (EMA) of the clustered latent features and codebook reset recipe was used to find inactivated codes during the training and reassign them according to input data.

\paragraph{Architecture.} Inspired by T2M-GPT~\cite{zhang2023generating}, we also adopt a simple convolutional architecture composed of 1D dilated convolution, residual block~\cite{he2016deep}, and ReLU. 
For our RVQ-VAE, we use 1D convolution with stride 2 and nearest interpolation for temporal downsampling and upsampling respectively. The downsampling rate is thus $l = 2^{L}$, where $L$ denotes the number of residual blocks. In our experiments, we set $L=3$ to obtain the downsampling rate $l=8$, each residual block then contains 2 dilated 1D convolution blocks, and the dilated factor is set to 3. More architecture details can be found in the supplementary material.

\paragraph{Discussion.} RVQ-VAE can make GPT-like models to generate high-quality motion with low computational costs effectively and push the upper bound of generation quality. For a fixed downsampling factor $l$, RVQ-VAE can produce more realistic reconstructions than VQ-VAE. RVQ can precisely approximate a latent feature using a given codebook size by increasing the coded depth number without extra computational burden. In addition, our experimental results demonstrate that the precise approximation by RQ-VAE allows a greater increase of $l$ and decrease of the latent discrete code time length $\lfloor T/l \rfloor$ than VQ-VAE, while achieving better reconstruction quality. Consequently, RVQ-VAE enables GPT-like models to reduce their computational costs, increase the speed of text-to-motion generation, and learn the long-range interactions of codes and textual features better.

\subsection{T2M-HiFiGPT}
\label{sec:t2m_hifigpt}
In this section, we propose T2M-HiFiGPT in~\cref{fig:pipeline}, to \textbf{autoregressively} predict a code stack of motion RVQ-VAE. After we obtain the residual codes extracted by RVQ-VAE. We introduce how our T2M-HiFiGPT efficiently learns the stacked map of discrete codes. We also propose our novel training strategy for T2M-HiFiGPT to prevent the exposure bias~\cite{ranzato2015sequence} problem during the training phase.

With a learned motion RVQ-VAE, a motion sequence $X=[x_{1},x_{2},\ldots,x_{T}]$ can be mapped to a 2D code matrix $S=\{s_{tj}|t=1,\ldots,\lfloor T/l \rfloor;j=1,\ldots,R;s_{ij} \in [0,K-1]\}$ of shape $\lfloor T/l \rfloor \times R$, where $S_{t}$ is the $t-th$ row of $S$, contains $R$ discrete codes written as
\begin{equation}
\label{eq:gpt_code}
S_{t}=(S_{t1},\ldots,S_{tR}),t=1,\ldots,\lfloor T/l \rfloor
\end{equation}
By projecting $S$ back to their corresponding codebook entries and sum along the residual axis according to \cref{eq:rvq_quant}, we obtain the quantized latent feature $Z^{\star}=[z_{1}^{\star},\ldots,z_{\lfloor T/l \rfloor}^{\star}]$, which can be decoded to a motion $X_{re}$ through the decoder $Dec$. Therefore, text-to-motion generation can be formulated as an autoregressive next-index prediction: given previous $t-1$ time indices and $w-1$ residual code sequences, and text condition $e_{text}$, we aim to predict the distribution of possible next indices $P(S_{ij}|e_{text};S_{t<i};S_{i,w<j})$, which can be addressed with a double-tier transformer structure similar to Valle~\cite{wang2023neural}. Note that our proposed T2M-HiFiGPT is a fully autoregressive model while Valle is not. Therefore, the probability distribution of 2D code matrix $S$ can be factorized autoregressively as follows:
\begin{equation}
    \label{eq:gpt_ar}
    P(S|e_{text})=\Pi_{t=1}^{\lfloor T/l \rfloor}\Pi_{w=1}^{R}P(S_{tw}|e_{text},S_{<t,w},S_{t,<w})
\end{equation}
Our optimization goal is to maximize the negative log-likelihood (NLL) of data distribution: 

\begin{equation}
    \label{eq:gpt_ar_nll_loss}
    L_{nll} = \sum_{t=1}^{\lfloor T/l \rfloor}\sum_{w=1}^{R}[-log(P(S_{tw}|e_{text};S_{<t,w},S_{t,<w}))]
\end{equation}
Since we must automatically infer the exact motion length corresponding to the input text, we must also maximize the log-likelihood of the binary stop status for each timestep, denoted as $Y={y_{1},y_{2},...y{\lfloor T/l \rfloor}}, y_{i} = 1$ if the current time step is the last frame otherwise 0. The loss function is formulated as:
\begin{equation}
    \label{eq:gpt_ar_stop_loss}
    L_{stop} = \sum_{t=1}^{\lfloor T/l \rfloor}[-logP(y_{t}|S_{\leq t};e_{text})]
\end{equation}
Therefore, the final loss is formulated as follows:
\begin{equation}
    \label{eq:gpt_loss}
    L = L_{nll} + \beta L_{stop} 
\end{equation}
In addition, we leverage CLIP~\cite{radford2021learning} to extract text embedding $e_{text}$, which has shown its effectiveness in relevant tasks~\cite{crowson2022vqgan,ramesh2022hierarchical,tevet2022motionclip,zhang2023generating,zhong2023attt2m}.

\subsubsection{Network Architecture}
A naiive method can unfold $S$ into a sequence of length $\lfloor T/l \rfloor \times R$ and feed it to the conventional transformer~\cite{vaswani2017attention}. However, it neither leverages the reduced length of $\lfloor T/l \rfloor$ by RVQ-VAE nor reduces the computational costs, In addition, it also neglects the temporal-residual relationship between the 2D discrete representations. Therefore, we propose T2M-HiFiGPT to efficiently learn the 2D temporal-residual codes extracted by RVQ-VAE with residual depth $R$. 

As shown in~\cref{fig:pipeline}, T2M-HiFiGPT consists of a temporal GPT and residual GPT. The temporal GPT is used to summarize the past frames from time 1 to current time $t$ and text embedding $e_{text}$ into a 1D context vector $F_{t}$, while the residual GPT regard $F_{t}$ as a prompt condition and leverage it to sequentially predict the next residual code. $F_{t}$ is also used to predict the stop token probability in \cref{eq:gpt_ar_stop_loss}. Both GPTs use standard transformers~\cite{vaswani2017attention} with causal attention masks and are jointly trained from end to end given the text CLIP embedding with the optimization loss function \cref{eq:gpt_loss}.

\paragraph{Temporal GPT.} The temporal GPT is a stack of transformer blocks with causal attention masks to extract a context vector that summarizes the information in previous frames and text embeddings. For the input of the temporal GPT, we reuse the learned codebook of motion RVQ-VAE with depth $R$, sum the residual features into the quantized latent feature, and add the corresponding absolute positional embedding. Specifically, we define the input $u_{t}$ of the temporal GPT as
\begin{equation}
\label{eq:gpt_temporal_input}
u_{t}=PET(t)+\sum_{j=1}^{R}C(S_{t,j}), t>1
\end{equation}
where $PET(t)$ is a learnable absolute positional embedding for temporal position
$t$. The text embedding $e_{text}$ is the input at the first position, which is also regarded as the start of a sequence. After the sequence $u_{t}$ is processed by the temporal GPT, a context vector $F_{t}$ encodes all information before time $t$ as 
\begin{equation}
    \label{eq:gpt_temporal_context}
    F_{t}=TemporalGPT(e_{text},u_{1},\ldots,u_{t})
\end{equation}
We also predict the binary stop status probability with a simple post layer norm and linear layer:
\begin{equation}
    \label{eq:gpt_temporal_stop}
    P(y_{t}) = sigmoid(Linear(LayerNorm(F_{t})))
\end{equation}
Note that at the inference phase, we always select the frame index with the maximum stop probability $P(y_{t}$ as the end frame.

\paragraph{Residual GPT.} Given the context vector $F_{t}$ as the prompt condition, the residual GPT autoregressively predicts $R$ codes $(S_{t1},\ldots,S_{tR})$ at timestep $t$. At timestep $t$ and depth $w$, the input $v_{tw}$ of the residual GPT is defined as the learned code embeddings such that
\begin{equation}
    \label{eq:gpt_residual_input}
    v_{tw} = PER(w) + C(S_{tw}), w \geq 1
\end{equation}
where $PER(w)$ is a learnable absolute positional embedding for depth $w$ and shared at every timestep $t$. The context vector $F_{t}$ is the input at the first position, which is also regarded as the start token. Accordingly, the residual GPT predicts the next code for a finer estimation of $S_{tw}$ based on the previous estimations $S_{t,<w}$. Finally, the residual GPT predicts the next token conditional distribution as follows:
\begin{equation}
    \label{eq:gpt_residual}
    P(S_{tw}|e_{text};S_{<t,w},S_{t,<w})=ResidualGPT(F_{t},v_{t,<w})
\end{equation}
which is used in the final NLL loss function in \cref{eq:gpt_ar_nll_loss}.

\paragraph{Discussion.} In comparison with the vallina GPT model proposed in T2M-GPT~\cite{zhang2023generating} and Att-T2M~\cite{zhong2023attt2m}, our proposed double-tier GPT structure can theoretically have a better computational advantage given the same number of transformer layers. The computational complexity of vallina GPT is $O(4 \times H \times \lfloor T/l \rfloor \times \lfloor T/l \rfloor)$, while our T2M-HiFiGPT is $O(H/2 \times \lfloor T/l \rfloor \times \lfloor T/l \rfloor + H/2 \times \lfloor T/l \rfloor \times R \times R)$, where $H$ denotes the total transformer layers, which is less than the vallina GPT model. In the following experimental section, we will compare the detailed parameter counts and inference speed with competitive GPT models~\cite{zhang2023generating,zhong2023attt2m}.

\subsection{Corrupted RVQ Augmentation and Condition Dropout}
The exposure bias~\cite{ranzato2015sequence} is known to deteriorate the performance of an auto-regressive (AR) model due to the error accumulation from the discrepancy of predictions in training and inference. During an inference of T2M-HiFiGPT, the prediction errors can also accumulate along with the depth $R$, since estimation of the finer feature vector becomes harder as the depth increases. To resolve this problem, we propose corrupted RVQ augmentation and a condition dropout strategy in our T2M-HiFiGPT training. 
\paragraph{Corrupted RVQ augmentation.} Similar to T2M-GPT~\cite{zhang2023generating}, we also find the positive gain from corrupted indices augmentation. However, since we adopt the RVQ-VAE framework, there is a stack of code indices in each timestep, thus the compressed motion sequence is a 2D matrix $\lfloor T/l \rfloor \times R$ with temporal-residual structure. Randomly replacing part of the discrete codes disregards the unique 2D structure and will lead to a noisy quantized input vector at each timestep due to our residual summation formula shown in \cref{eq:rvq_quant}, therefore, the temporal GPT will learn nothing except corrupted signals. Accordingly, we must replace $\tau \times 100\%$ ($\tau=0.5$ in our experiment) ground-truth code indices by the timestep granularity,i.e., replace each row with random code indices rather than all elements in the 2D residual indices matrix (per-code corruption). Note that this strategy will keep ground truth indices at $\tau \times 100\%$ \textbf{timesteps}. We will provide an ablation study on this strategy in our experiment section to demonstrate its importance.
\paragraph{Condition dropout strategy.} To learn the accurate relation between low-level and high-level codes, we can randomly drop the CLIP condition vector $e_{text}$ and replace it with a zero vector, equal to the \textit{NULL} prompt condition, in the same spirit as the T2I diffusion model training~\cite{ramesh2021zero}. Specifically, during the training phase, we randomly replace the CLIP condition vector $e_{text}$ with zero vectors:
\begin{equation}
    \label{eq:corrupt_dropout}
     \hat{e}_{text}=
    \begin{cases}
    e_{text} & \text{Rand} > p_{drop} \\
    \textbf{0} & \text{otherwise}
    \end{cases}
\end{equation}

Therefore, we input $\hat{e}_{text}$ into the residual GPT. The $p_{drop}$ is set to 0.1 experimentally. During the inference phase, we adopt a similar classifier-free guidance (CFG)~\cite{ho2022classifier} strategy (originally used in the diffusion model) in our T2M-HiFiGPT to boost the text-motion matching accuracy, that is, for \textit{each} discrete code at \textit{each} time step, we mix their conditional $G_{\text{cond}}$ and unconditional logits $G_{\text{uncond}}$ with certain fixed guidance scale parameter $\gamma$, and leverage the mixed logits $G_{\text{mix}}$ for the multi-norminal sampling:
\begin{equation}
    \label{eq:corrupt_cfg}
    G_{\text{mix}} = G_{\text{uncond}} + \gamma(G_{\text{cond}} - G_{\text{uncond}})
\end{equation}
Note that the unconditional logits are obtained by replacing each input CLIP vector $e_{text}$ with zero vector and forward through the double-tier GPT described above, we conduct CFG on both discrete motion code and stop status logits. In our ablation experiments, we have found the conditional dropout training and classifier-free guidance inference strategy can indeed have a very positive gain in the objective metric and act as an important regularizer in our proposed double-tier GPT structure. 

\section{Experiments}
\label{sec:exp}
In this section, we present our experimental results. In section 4.1 and 4.2, we introduce standard datasets as well as evaluation metrics. In section 4.3, we describe the implementation details of T2M-HiFiGPT. We compare our results to competitive approaches in Section 4.4. Finally, we provide analysis and discussion in Section 4.5.
\subsection{Dataset Details}
We evaluate our approach using two datasets for text-driven motion generation: HumanML3D~\cite{guo2022generating} and KIT Motion-Language (KIT-ML)~\cite{plappert2016kit}. These datasets are widely recognized and utilized within the research community. KIT-ML~\cite{plappert2016kit} comprises 3,911 human motion sequences paired with 6,278 textual annotations. The dataset features a vocabulary of 1,623 unique words, excluding variations in capitalization and punctuation. The motion sequences, sourced from the KIT~\cite{mandery2015kit} and CMU~\cite{cmu_mocap} datasets, are downsampled to 12.5 frames per second (FPS). Each sequence is accompanied by one to four descriptive sentences, with an average description length of about eight words. Following established protocols, we divide the dataset into training, validation, and test sets in an 80\%, 5\%, and 15\% ratio, respectively. The model that performs best on the validation set, as measured by the Fréchet Inception Distance (FID), is then evaluated on the test set. HumanML3D~\cite{guo2022generating}, currently the most extensive 3D human motion dataset with textual annotations, contains 14,616 human motions and 44,970 text descriptions. The text corpus encompasses 5,371 distinct words. The motion sequences, derived from AMASS~\cite{mahmood2019amass} and HumanAct12~\cite{guo2020action2motion}, undergo specific preprocessing: they are scaled to 20 FPS, motions exceeding 10 seconds are randomly cropped to that duration, re-targeted to a standard human skeletal template, and oriented to initially face the Z+ direction. Each motion is associated with at least three detailed textual descriptions, averaging around 12 words in length. Consistent with prior work~\cite{guo2022tm2t}, we partition this dataset into training, validation, and test sets with the same 80\%, 5\%, and 15\% distribution. Performance on the test set is reported for the model with the best FID score on the validation set. Note that we use the same motion representation as the prior work~\cite{zhang2023generating}.
\subsection{Evaluation Metrics}
Following~\cite{guo2022generating}, global representations of motion and text descriptions are first extracted with the pre-trained network in~\cite{guo2022generating}, and then measured by the following five metrics:
\begin{itemize}
    \item \textbf{R-Precision.} We are provided with a single motion sequence and a set of 32 text descriptions, which includes one ground-truth description and 31 mismatched descriptions chosen at random. We compute the Euclidean distances between the embeddings of the motion and each text description. The retrieval performance is then evaluated by reporting the Top-1, Top-2, and Top-3 accuracy rates, which reflect the model's ability to correctly match the motion sequence with its corresponding ground-truth text description among the top-ranked results.
    \item \textbf{Frechet Inception Distance (FID).} We employ the Fréchet Inception Distance (FID) metric, which measures the distribution distance between the two sets of motion data based on extracted motion features. This metric is widely used to evaluate the quality of generative models by comparing the statistical distribution of generated data to that of real data.
    \item \textbf{Multimodal Distance (MM-Dist).} The average Euclidean distances between each text feature and the generated motion feature.
    \item \textbf{Diversity.} we randomly select 300 motion pairs from the set, then extract features from these motions and calculate the average Euclidean distance between each pair. This average distance serves as a quantitative measure of the diversity present within the set of motions.
    \item \textbf{Multimodality (MModality).} We generate 20 distinct motion sequences, which we then organize into 10 unique pairs. For each pair, we extract motion features and calculate the average Euclidean distance between them. This process is repeated for all text descriptions, and the resulting average distances are aggregated. The final reported metric is the overall average of these distances, providing a measure of the variation among the generated motions corresponding to the same textual prompt.
\end{itemize}

\subsection{Implementation Details }
For Motion RVQ-VAE, considering the trade-off between computational time and synthesis accuracy, we set the codebook size to 256 × 512, i.e., 256 512-dimension dictionary vectors. The downsampling rate $l$ is 8 rather than 4 in T2M-GPT~\cite{zhang2023generating}. We provide an ablation on the number of residual depth and codebook sizes in Section 4.5. For both the HumanML3D~\cite{guo2022generating} and KIT-ML~\cite{plappert2016kit} datasets, the motion sequences are cropped to $T = 32$ for training. We use AdamW~\cite{loshchilov2017decoupled} optimizer with $\beta_{1}=0.9,\beta_{2}=0.95$, the training batch size is 256. We train the first 250 epochs
with a learning rate of 2e-4, and then slowly decay the learning rate with a cosine scheduler for the rest 3350 epochs. $\alpha_{1,2}=0,5,\alpha_{3}=1.0$ in the objective function~\cref{eq:rvq_recon}, respectively. Following~\cite{guo2022generating,zhang2023generating}, the datasets KIT-ML~\cite{plappert2016kit} and HumanML3D~\cite{guo2022generating} are extracted into motion features with dimensions 251 and 263
respectively, which correspond to local joint position, velocity, and rotations in root space as well as global translation and rotations. These features are computed from 21 and 22 joints of SMPL~\cite{loper2023smpl}. We train the RVQ-VAE on a single V100 card. 
For the T2M-HiFiGPT, we employ 18 transformer~\cite{vaswani2017attention} layers with a dimension of 512 and 16 heads, 9 layers for the temporal GPT and 9 layers for the residual GPT respectively. Following Guo et al.~\cite{guo2022generating}, the maximum length of motion is 196 on both datasets, and the minimum lengths are 40 and 24 for HumanML3D and KIT-ML respectively. The maximum length of the code index sequence is $\lfloor T/l \rfloor =24$. The transformer is optimized also using AdamW~\cite{loshchilov2017decoupled} with $\beta_{1}=0.9,\beta_{2}=0.95$ and the batch size is 128. The initialized learning rate is set to 1e-4 for 200 epochs and decays slowly with a cosine scheduler for another 1400 epochs. We train T2M-HiFiGPT on 8 V100 cards for faster speed. More details can be found in the supplementary material.

\begin{table}
\centering
\caption{Compare the parameters and computational cost against relevant competitive GPT-based methods at the \textit{reconstruction} stage. Time/Frame(ms) denotes the average time cost when generating a motion frame. We measure those metrics by recording the detailed time cost during our testing on the HumanML3D test set. Our code is implemented using Pytorch.}\vspace{5pt}
\scalebox{0.8}{
\begin{tabular}{ccc}
\toprule
Method & \#Parameter(M) & Time/Frame(ms) \\
\midrule
T2M-GPT~\cite{zhang2023generating} & \textbf{19.7} & 0.016 \\
\midrule
Att-T2M~\cite{zhong2023attt2m} & 20.49 & 0.017 \\
\midrule
Ours & 21.9 & \textbf{0.015} \\
\bottomrule
\end{tabular}
}
\label{tab:compare_time_reconstruction}
\end{table}

\begin{table}
\centering
\caption{Compare the parameters and computational cost against relevant competitive GPT-based methods at the \textit{generation} stage. Time/Frame(ms) denotes the average time cost when generating a motion frame. Time/Token(ms) denotes the average time cost when generating a single discrete motion token by GPT. Both indicators serve as metrics for the response time, and as illustrated in the table, all methods are capable of achieving real-time output at high frame rates.}\vspace{5pt}
\resizebox{\linewidth}{!}{
 \setlength{\tabcolsep}{3.5pt}
\begin{tabular}{cccc}
\toprule
Method & \#Parameter(M) & Time/Token(ms) & Time/Frame(ms)\\
\midrule
T2M-GPT~\cite{zhang2023generating} & 228 & 13.1 & \textbf{3.2} \\
\midrule
Att-T2M~\cite{zhong2023attt2m} & 254 & 14.5 & 3.6 \\
\midrule
Ours & \textbf{58} & \textbf{6.4} & 7.8\\
\bottomrule
\end{tabular}
}
\label{tab:compare_time_generation}
\end{table}

\begin{table*}
\centering
\caption{\textbf{Comparison with the state-of-the-art methods on HumanML3D~\cite{guo2022generating} test set.} We compute standard metrics following Guo et al.~\cite{guo2022generating}. For each metric, we repeat the evaluation 20 times and report the average with 95\% confidence interval. The bold text indicates the best result and the second best is underlined.}\vspace{5pt}

\begin{tabular}{cccccccc}
        \toprule
        \multirow{2}*{Methods} & \multicolumn{3}{c}{R Precision $\uparrow$} & \multirow{2}*{FID$\downarrow$} & \multirow{2}*{MM-D$\downarrow$} & \multirow{2}*{Div$\rightarrow$} & \multirow{2}*{MM$\uparrow$} \\
        \cline{2-4}
        ~ & Top-1 & Top-2 & Top-3 & ~ & ~ & ~ & ~ \\
        \midrule
        Real & $0.511_{\pm.003}$ & $0.703_{\pm.003}$ & $0.797_{\pm.002}$ & $0.002_{\pm.000}$ & $2.974_{\pm.008}$ & $9.503_{\pm.065}$ & - \\
        T2M-GPT~\cite{zhang2023generating} (Recons,) & $0.493_{\pm.003}$ & $0.685_{\pm.002}$ & $0.782_{\pm.002}$ & $0.09_{\pm.001}$ & $3.093_{\pm.009}$ & $9.676_{\pm.069}$ & - \\
        Att-T2M~\cite{zhong2023attt2m} (Recons,) & $0.498_{\pm.002}$ & $0.690_{\pm.002}$ & $0.784_{\pm.002}$ & $0.109_{\pm.001}$ & $3.071_{\pm.007}$ & $9.699_{\pm.089}$ & - \\
        Our RVQ-VAE (Recons.) & $0.510_{\pm.002}$ & $0.703_{\pm.002}$ & $0.796_{\pm.002}$ & $0.027_{\pm.000}$ & $2.995_{\pm.006}$ & $9.591_{\pm.085}$ & - \\
        \midrule
        Seq2Seq~\cite{lin2018generating} & $0.180_{\pm.002}$ & $0.300_{\pm.002}$ & $0.396_{\pm.002}$ & $11.75_{\pm.035}$ & $5.529_{\pm.007}$ & $6.223_{\pm.061}$ & - \\
        Language2Pose~\cite{ahuja2019language2pose} & $0.246_{\pm.002}$ & $0.387_{\pm.002}$ & $0.486_{\pm.002}$ & $11.02_{\pm.046}$ & $5.296_{\pm.008}$ & $7.676_{\pm.058}$ & - \\
        Text2Gesture~\cite{bhattacharya2021text2gestures} & $0.165_{\pm.001}$ & $0.267_{\pm.002}$ & $0.345_{\pm.002}$ & $5.012_{\pm.030}$ & $6.030_{\pm.008}$ & $6.409_{\pm.071}$ & - \\
        Hier~\cite{ghosh2021synthesis} & $0.301_{\pm.002}$ & $0.425_{\pm.002}$ & $0.552_{\pm.004}$ & $6.532_{\pm.024}$ & $5.012_{\pm.018}$ & $8.332_{\pm.042}$ & - \\
        MoCoGAN~\cite{tulyakov2018mocogan} & $0.037_{\pm.000}$ & $0.072_{\pm.001}$ & $0.106_{\pm.001}$ & $94.41_{\pm.021}$ & $9.643_{\pm.006}$ & $0.462_{\pm.008}$ & $0.019_{\pm.000}$ \\
        Dance2Music~\cite{lee2019dancing} & $0.033_{\pm.000}$ & $0.065_{\pm.001}$ & $0.097_{\pm.001}$ & $66.98_{\pm.016}$ & $8.116_{\pm.006}$ & $0.725_{\pm.011}$ & $0.043_{\pm.001}$ \\
        TM2T~\cite{guo2022tm2t}  & $0.424_{\pm.003}$ & $0.618_{\pm.003}$ & $0.729_{\pm.002}$ & $1.501_{\pm.017}$ & $3.467_{\pm.011}$ & $8.589_{\pm.076}$ & $2.424_{\pm.093}$ \\
        Guo et al.~\cite{guo2022generating} & $0.455_{\pm.003}$ & $0.636_{\pm.003}$ & $0.736_{\pm.002}$ & $1.087_{\pm.021}$ & $3.347_{\pm.008}$ & $9.175_{\pm.083}$ & $2.219_{\pm.074}$ \\
        MDM~\cite{tevet2022human} & - & - & $0.611_{\pm.007}$ & $0.544_{\pm.044}$ & $5.566_{\pm.027}$ & $\textbf{9.559}_{\pm.086}$ & $\textbf{2.799}_{\pm.072}$ \\
        MotionDiffuse~\cite{zhang2022motiondiffuse} & $0.491_{\pm.001}$ & $0.681_{\pm.001}$ & $0.782_{\pm.001}$ & $0.630_{\pm.001}$ & $3.113_{\pm.001}$ & $\underline{9.410}_{\pm.049}$ & $1.553_{\pm.042}$ \\
        MLD~\cite{chen2023executing} & $0.481_{\pm.003}$ & $0.673_{\pm.003}$ & $0.772_{\pm.002}$ & $0.473_{\pm.013}$ & $3.196_{\pm.010}$ & $9.724_{\pm.082}$ & $2.413_{\pm.079}$ \\
        T2M-GPT~\cite{zhang2023generating} & $0.491_{\pm.003}$ & $0.680_{\pm.003}$ & $0.775_{\pm.002}$ & $0.116_{\pm.004}$ & $3.118_{\pm.011}$ & $9.761_{\pm.081}$ & $1.856_{\pm.011}$ \\
        Att-T2M~\cite{zhong2023attt2m} & $0.499_{\pm.003}$ & $0.690_{\pm.002}$ & $0.786_{\pm.002}$ & $0.112_{\pm.006}$ & $3.038_{\pm.007}$ & $9.700_{\pm.090}$ & $2.452_{\pm.051}$ \\
        ReMoDiffuse~\cite{zhang2023remodiffuse} & $\underline{0.510}_{\pm.002}$ & $\underline{0.69}8_{\pm.006}$ & $\underline{0.795}_{\pm.004}$ & $\underline{0.103}_{\pm.004}$ & $\underline{2.974}_{\pm.016}$ & $9.018_{\pm.075}$ & $1.795_{\pm.043}$ \\
        FineMoGen~\cite{zhang2023finemogen} & $0.504_{\pm.002}$ & $0.690_{\pm.002}$ & $0.784_{\pm.002}$ & $0.151_{\pm.008}$ & $2.998_{\pm.008}$ & $9.263_{\pm.094}$ & $\underline{2.696}_{\pm.079}$ \\
        \midrule
        Ours & $\textbf{0.514}_{\pm.003}$ & $\textbf{0.708}_{\pm.003}$ & $\textbf{0.802}_{\pm.002}$ & $\textbf{0.066}_{\pm.004}$ & $\textbf{2.940}_{\pm.009}$ & $9.718_{\pm.099}$ & $1.502_{\pm.041}$ \\
    \bottomrule
\end{tabular}
\vspace{6pt}
\label{table:sota_hm3d}
\end{table*}

\begin{table*}
\centering
\caption{\textbf{Comparison with the state-of-the-art methods on KIT~\cite{plappert2016kit} test set.} The experimental settings are the same as~\cref{table:sota_hm3d}. We report the metrics following T2M~\cite{guo2022generating} and repeat 20 times to get the average results with 95\% confidence interval. The best results are marked in bold and the second best is underlined.} \vspace{5pt}
\begin{tabular}{cccccccc}
        \toprule
        \multirow{2}*{Methods} & \multicolumn{3}{c}{R Precision $\uparrow$} & \multirow{2}*{FID$\downarrow$} & \multirow{2}*{MM-D$\downarrow$} & \multirow{2}*{Div$\rightarrow$} & \multirow{2}*{MM$\uparrow$} \\
        \cline{2-4}
        ~ & Top-1 & Top-2 & Top-3 & ~ & ~ & ~ & ~ \\
        \midrule
        Real & $0.424_{\pm.005}$ & $0.649_{\pm.006}$ & $0.779_{\pm.006}$ & $0.031_{\pm.004}$ & $2.788_{\pm.012}$ & $11.08_{\pm.097}$ & - \\
        Our RVQ-VAE (Recons.) & $0.414_{\pm.003}$ & $0.625_{\pm.004}$ & $0.747_{\pm.004}$ & $0.154_{\pm.003}$ & $2.860_{\pm.014}$ & $10.951_{\pm.113}$ & - \\
        T2M-GPT~\cite{zhang2023generating} (Recons.) & $0.399_{\pm.005}$ & $0.614_{\pm.005}$ & $0.740_{\pm.006}$ & $0.472_{\pm.011}$ & $2.986_{\pm.027}$ & $10.994_{\pm.120}$ & - \\
        Att-T2M~\cite{zhong2023attt2m} (Recons.) & $0.391_{\pm.005}$ & $0.604_{\pm.007}$ & $0.733_{\pm.006}$ & $0.562_{\pm.014}$ & $3.025_{\pm.019}$ & $11.088_{\pm.093}$ & - \\
        \midrule
        Seq2Seq~\cite{lin2018generating} & $0.103_{\pm.003}$ & $0.178_{\pm.005}$ & $0.241_{\pm.006}$ & $24.86_{\pm.348}$ & $7.960_{\pm.031}$ & $6.744_{\pm.106}$ & - \\
        Language2Pose~\cite{ahuja2019language2pose} & $0.221_{\pm.005}$ & $0.373_{\pm.004}$ & $0.483_{\pm.005}$ & $6.545_{\pm.072}$ & $5.147_{\pm.030}$ & $9.073_{\pm.100}$ & - \\
        Text2Gesture~\cite{bhattacharya2021text2gestures} & $0.156_{\pm.004}$ & $0.255_{\pm.004}$ & $0.338_{\pm.005}$ & $12.12_{\pm.183}$ & $6.964_{\pm.029}$ & $9.334_{\pm.079}$ & - \\
        Hier~\cite{ghosh2021synthesis} & $0.255_{\pm.006}$ & $0.432_{\pm.007}$ & $0.531_{\pm.007}$ & $5.203_{\pm.107}$ & $4.986_{\pm.027}$ & $9.563_{\pm.072}$ & - \\
        MoCoGAN~\cite{tulyakov2018mocogan} & $0.022_{\pm.002}$ & $0.042_{\pm.003}$ & $0.063_{\pm.003}$ & $82.69_{\pm.242}$ & $10.47_{\pm.012}$ & $3.091_{\pm.043}$ & $0.250_{\pm.009}$ \\
        Dance2Music~\cite{lee2019dancing} & $0.031_{\pm.002}$ & $0.058_{\pm.002}$ & $0.086_{\pm.003}$ & $115.4_{\pm.240}$ & $10.40_{\pm.016}$ & $0.241_{\pm.004}$ & $0.062_{\pm.002}$ \\
        TM2T~\cite{guo2022tm2t}  & $0.280_{\pm.005}$ & $0.463_{\pm.006}$ & $0.587_{\pm.005}$ & $3.599_{\pm.153}$ & $4.591_{\pm.026}$ & $9.473_{\pm.117}$ & $\textbf{3.292}_{\pm.081}$ \\
        Guo et al.~\cite{guo2022generating} & $0.361_{\pm.006}$ & $0.559_{\pm.007}$ & $0.681_{\pm.007}$ & $3.022_{\pm.107}$ & $3.488_{\pm.028}$ & $10.72_{\pm.145}$ & $2.052_{\pm.107}$ \\
        MDM~\cite{tevet2022human} & - & - & $0.396_{\pm.004}$ & $0.497_{\pm.021}$ & $9.191_{\pm.022}$ & $10.847_{\pm.109}$ & $1.907_{\pm.214}$ \\
        MotionDiffuse~\cite{zhang2022motiondiffuse} & $0.417_{\pm.004}$ & $0.621_{\pm.004}$ & $0.739_{\pm.004}$ & $1.954_{\pm.062}$ & $2.958_{\pm.005}$ & $\textbf{11.10}_{\pm.143}$ & $0.730_{\pm.013}$ \\
        MLD~\cite{chen2023executing} & $0.390_{\pm.008}$ & $0.609_{\pm.008}$ & $0.734_{\pm.007}$ & $0.404_{\pm.027}$ & $3.204_{\pm.027}$ & $10.80_{\pm.117}$ & $2.192_{\pm.071}$ \\
        T2M-GPT~\cite{zhang2023generating} & $0.402_{\pm.006}$ & $0.619_{\pm.005}$ & $0.737_{\pm.006}$ & $0.717_{\pm.041}$ & $3.053_{\pm.026}$ & $10.86_{\pm.094}$ & $1.912_{\pm.036}$ \\
        Att-T2M~\cite{zhong2023attt2m} & $0.413_{\pm.006}$ & $0.632_{\pm.006}$ & $0.751_{\pm.006}$ & $0.870_{\pm.039}$ & $3.039_{\pm.021}$ & $10.96_{\pm.123}$ & $\underline{2.281}_{\pm.047}$ \\
        ReMoDiffuse~\cite{zhang2023remodiffuse} & $0.427_{\pm.014}$ & $0.641_{\pm.004}$ & $0.765_{\pm.055}$ & $0.155_{\pm.006}$ & $2.814_{\pm.012}$ & $10.80_{\pm.105}$ & $1.239_{\pm.028}$ \\
        FineMoGen~\cite{zhang2023finemogen} & $0.432_{\pm.006}$ & $0.649_{\pm.005}$ & $0.772_{\pm.006}$ & $\underline{0.178}_{\pm.007}$ & $2.869_{\pm.014}$ & $10.85_{\pm.115}$ & $1.877_{\pm.093}$ \\
        \midrule
        Ours  & $\underline{0.428}_{\pm.006}$ & $\textbf{0.650}_{\pm.005}$ & $\textbf{0.780}_{\pm.006}$ & $0.337_{\pm.012}$ & $\textbf{2.777}_{\pm.019}$ & $\underline{11.19}_{\pm.099}$ & $1.434_{\pm.031}$ \\
    \bottomrule
\end{tabular}
\vspace{-12pt}
\label{table:sota_kit}
\end{table*}


\subsection{Comparison to state-of-the-art approaches}
We compare our T2M-HiFiGPT to existing state-of-the-art methods~\cite{ahuja2019language2pose,bhattacharya2021text2gestures,ghosh2021synthesis,guo2022generating,guo2022tm2t,lee2019dancing,lin2018generating,tevet2022human,tulyakov2018mocogan,zhang2022motiondiffuse,zhang2023generating,zhong2023attt2m,chen2023executing} on the test set of HumanML3D~\cite{guo2022generating} and KIT-ML~\cite{plappert2016kit}. Note that more visual results are provided within the supplementary material.

\paragraph{Quantitative results.} We show the comparison results in
~\cref{table:sota_hm3d} and~\cref{table:sota_kit} on HumanML3D~\cite{guo2022generating} test set and KIT-ML~\cite{plappert2016kit} test set. From~\cref{table:sota_hm3d}, on both datasets, our reconstruction with RVQ-VAE reaches close performances to real motion and achieves better construction quality than the recent VQ-based methods, which suggests high-quality discrete representations learned by our RVQ-VAE. For the generation, our approach achieves state-of-the-art performance on the FID metric compared to the state-of-the-art GPT-based two-stage method Att-T2M~\cite{zhong2023attt2m} and diffusion-based method ReMoDiffuse~\cite{zhang2023remodiffuse} or FineMoGen~\cite{zhang2023finemogen}, while significantly surpassing them with the matching distance metric and recall precision. KIT-ML~\cite{plappert2016kit} and HumanML3D~\cite{guo2022generating} are on different scales, which demonstrates the robustness of the proposed approach. Manually corrupting sequences and randomly dropping CLIP condition vectors during the training of double-tier GPT brings consistent improvement. A more detailed analysis is provided in the next section. Note that our framework can both automatically predict the generated motion and its length from the binary stop logits~\cref{eq:gpt_ar_stop_loss} in an online manner, which is very intuitive and demonstrated to be practical. In contrast, the diffusion-based methods such as MDM~\cite{tevet2022human}, FineMoGen~\cite{zhang2023finemogen}, ReMoDiffuse~\cite{zhang2023remodiffuse}, MotionDiffuse~\cite{zhang2022motiondiffuse} and MLD~\cite{chen2023executing} evaluate their models with the ground-truth motion length; their methods \textit{cannot} automatically generate accurate motion length, which is very important for real applications.
\paragraph{Computational Cost and Network Parameter Quantity.} We compare the computational time and network parameter quantity between our method and the most relevant GPT-based two-stage methods, T2M-GPT~\cite{zhang2023generating} and Att-T2M~\cite{zhong2023attt2m} in~\cref{tab:compare_time_reconstruction} and~\cref{tab:compare_time_generation}. From~\cref{tab:compare_time_reconstruction} and~\cref{tab:compare_time_generation}, we can observe that at the reconstruction stage, all methods have almost the same parameter numbers. However, our RVQ-VAE has faster inference speed than the two VQ-VAE-based methods, while achieving optimal reconstruction performance that is very close to the ground truth motion, confirming the effectiveness of our RVQ-VAE in the representation of human motion data. At the generation stage, our proposed double-tier GPT has four times fewer parameters and comparable inference speed while achieving SOTA performance in the text-to-motion task, which indicates our parameter efficacy and competitive computational cost in a practical system. Accordingly, our framework can both represent high-fidelity motion data and generate high-quality motion with textual description at a fast speed with fewer parameters.

\begin{figure*}[t]
 \centering
\includegraphics[width=\textwidth]{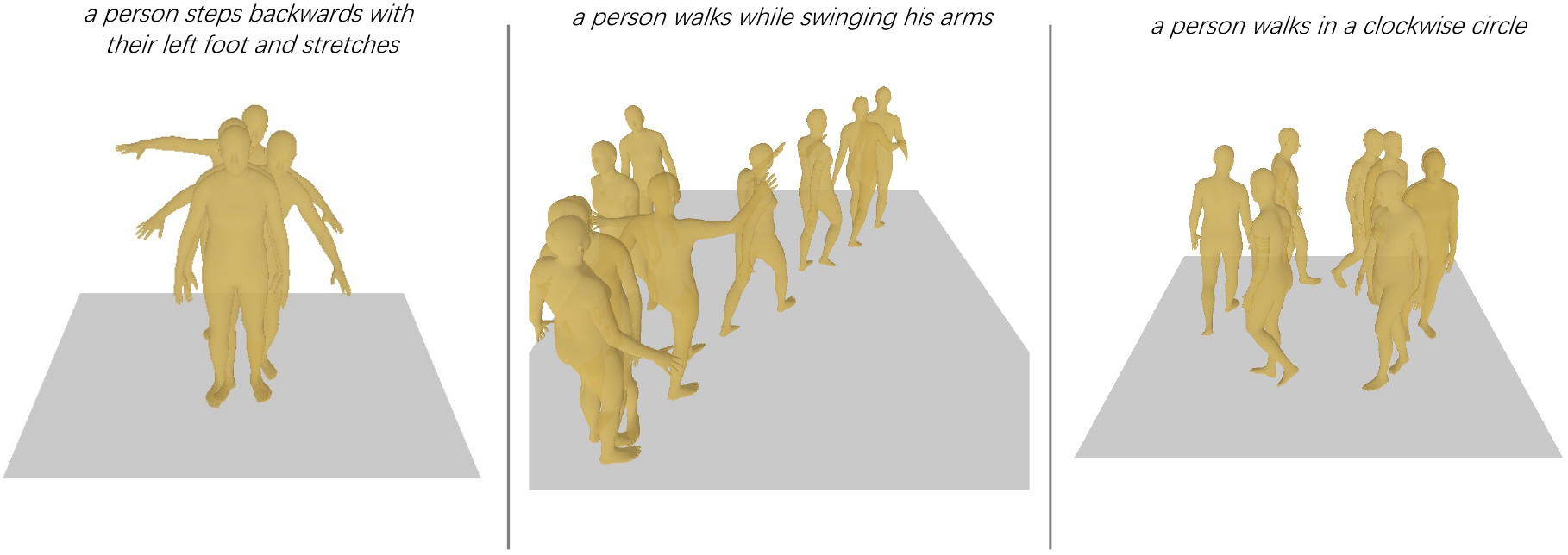}
\caption{\textbf{Visual results on HumanML3D~\cite{guo2022generating}.} Our T2M-HiFiGPT can generate precise and high-fidelity human motion consistent with the given text descriptions.}
\label{fig:teaser}
\end{figure*}

\begin{figure}[t]
 \centering
\includegraphics[width=\columnwidth]{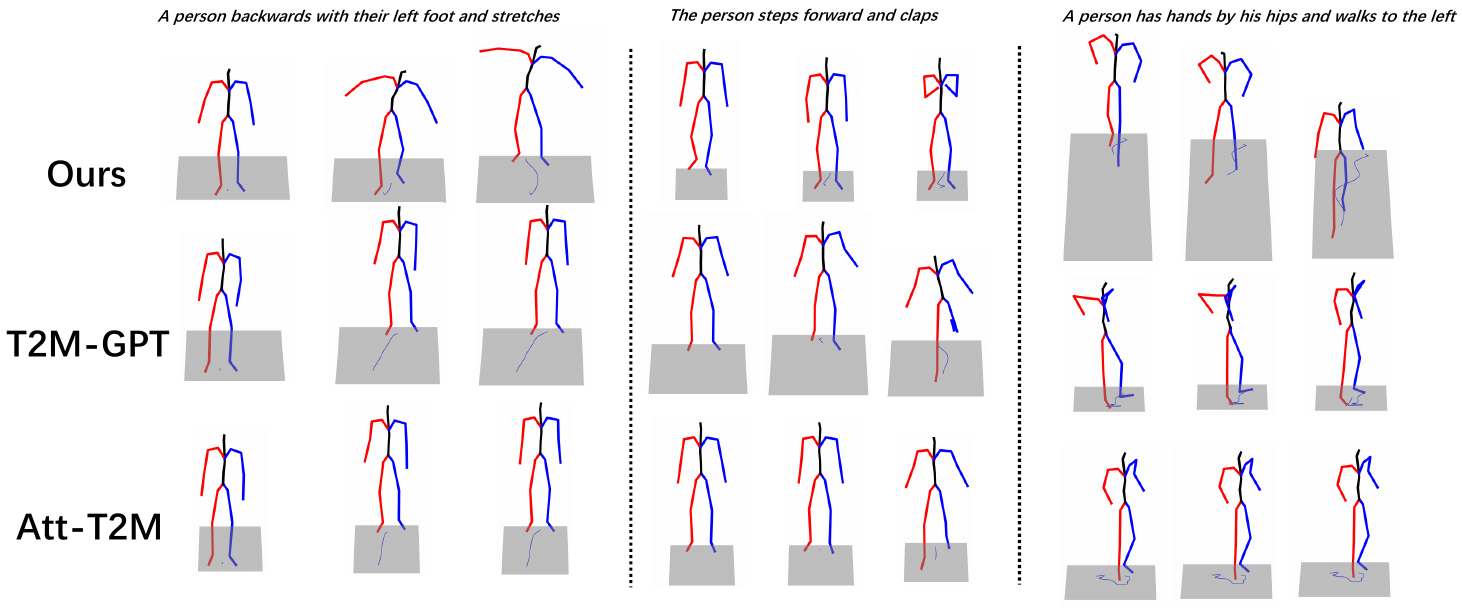}
\caption{\textbf{Qualitative comparison.} We compare our method qualitatively with Att-T2M~\cite{zhong2023attt2m} and T2M-GPT~\cite{zhang2023generating}, both adopting the GPT structure and achieving the most similar results as ours in the quantitative evaluation. Remarkably, our results more precisely match the textual descriptions. For a more
intuitive comparison, we demonstrate skeleton sequences and root trajectories here.}
\label{fig:visual_compare}
\end{figure}

\paragraph{Qualitative comparison.} ~\cref{fig:teaser} shows some visual results on HumanML3D~\cite{guo2022generating}. We compare our generations with the current GPT-based state-of-the-art models: T2M-GPT~\cite{zhang2023generating}, Att-T2M~\cite{zhong2023attt2m} in~\cref{fig:visual_compare}. From the examples in~\cref{fig:visual_compare}, one can figure out that our model generates human motion with better text-to-motion consistency than the others. For example, the generated motions of Att-T2M~\cite{tevet2022human} and T2M-GPT~\cite{zhang2023generating} either neglect some key actions such as "stretches" and "claps" or confuse the locomotion direction. Accordingly, they are not related to the semantics of the description. Note that more visualization results are provided in the supplementary material.

\subsection{Ablation study}
We conducted the following ablation studies on the HumanML3D~\cite{guo2022generating} training and test set due to its largest scale and diversity.
\paragraph{Evaluation on the importance of bone and velocity loss functions.} The velocity and acceleration loss functions can effectively ensure the dynamic characteristics of motion data. Moreover, the bone loss adopted in our RVQ-VAE reconstruction also plays an important role in achieving optimal performance. The bone loss function contains domain knowledge of the human skeleton structure and its intrinsic rigid relationship, thus boosting reconstruction accuracy. To verify their importance, we leave out one of them and observe the reconstruction performance. From ~\cref{tab:ab_geoloss}, we can clearly observe that, when both loss functions are removed, the performance is inferior. When both losses are utilized, the text-to-motion matching performance is optimal. This finding suggests that we can easily embed domain priors into our framework by simply inserting the domain-related prior loss into our RVQ-VAE, and then the generated discrete codes can implicitly comprise the domain priors. 

\begin{table}
\centering
\caption{Evaluation on the importance of the  bone and velocity loss functions in the proposed motion RVQ-VAE.}\vspace{5pt}
 \adjustbox{width=\linewidth}{
 \setlength{\tabcolsep}{3.5pt}
\begin{tabular}{cc|cccc}
\toprule
 Velocity Loss & Bone Loss & FID$\downarrow$ & MM-Dist$\downarrow$ & Top-1$\uparrow$ & Top-3$\uparrow$\\
\midrule
\xmark & \xmark & 0.030 & 3.014 & 0.508 & 0.791\\
\midrule
\cmark & \xmark &\textbf{0.024} & 3.001 & \textbf{0.510} & 0.793\\
\midrule
\cmark & \cmark & 0.027 & \textbf{2.995} & \textbf{0.510} & \textbf{0.796} \\
\bottomrule
\end{tabular}
}
\label{tab:ab_geoloss}
\end{table}

\begin{figure}[t]
 \centering
 \begin{subfigure}[b]{0.475\columnwidth}
     \centering
     \includegraphics[width=\textwidth]{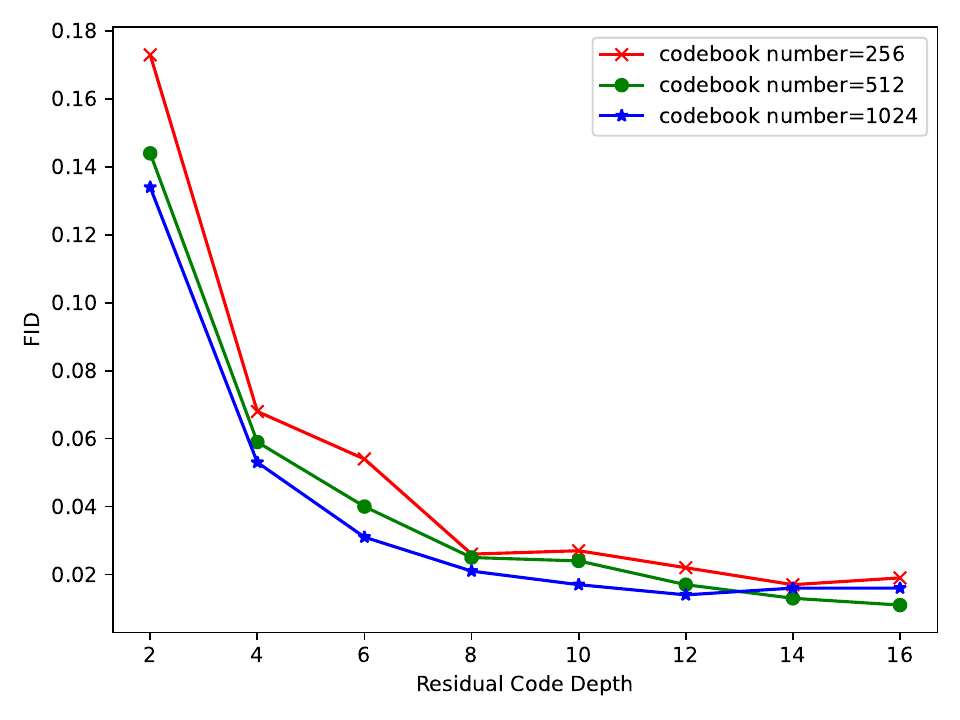}
     \caption{\textbf{FID}}
 \end{subfigure}
 \hfill
 \begin{subfigure}[b]{0.475\columnwidth}
     \centering
     \includegraphics[width=\textwidth]{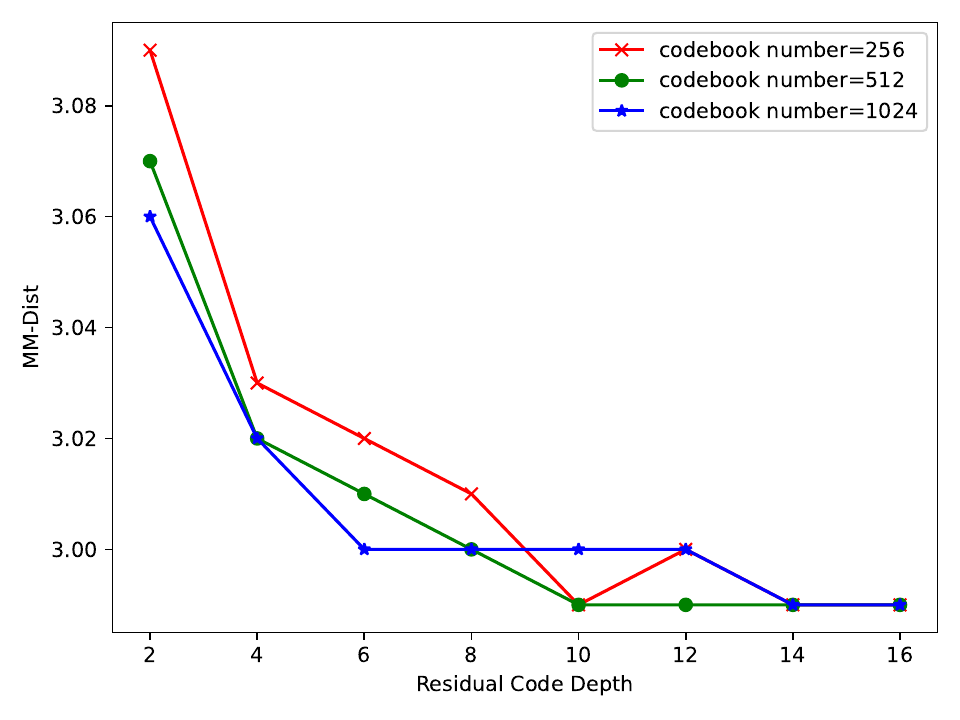}
     \caption{\textbf{MM-Dist}}
 \end{subfigure}
 \hfill
 \begin{subfigure}[b]{0.475\columnwidth}
     \centering
     \includegraphics[width=\textwidth]{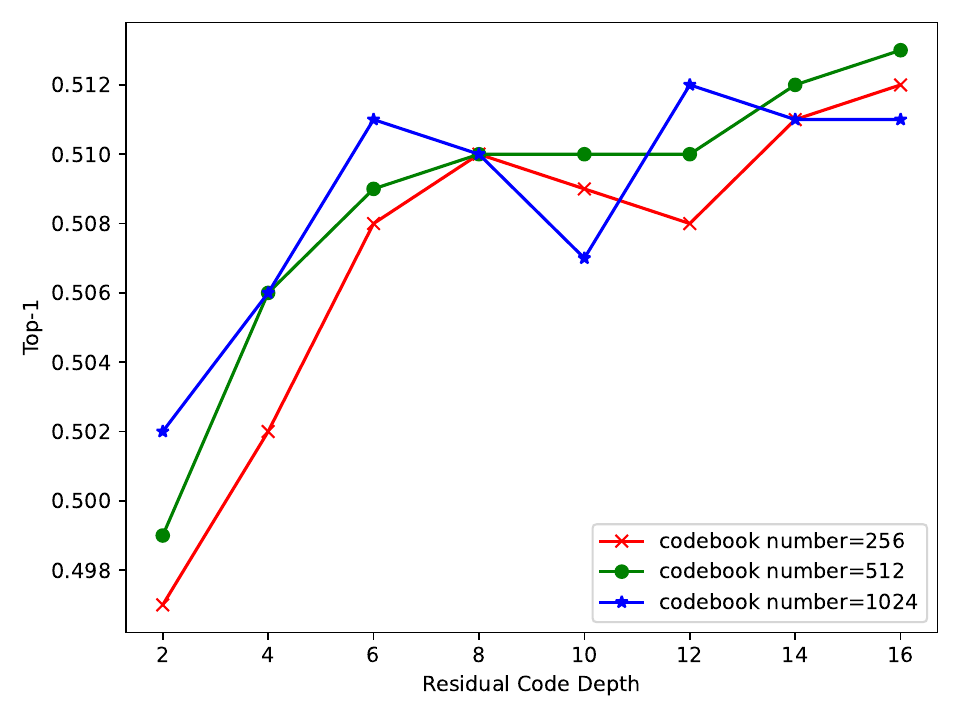}
     \caption{\textbf{Top-1 accuracy}}
 \end{subfigure}
 \hfill
 \begin{subfigure}[b]{0.475\columnwidth}
     \centering
     \includegraphics[width=\textwidth]{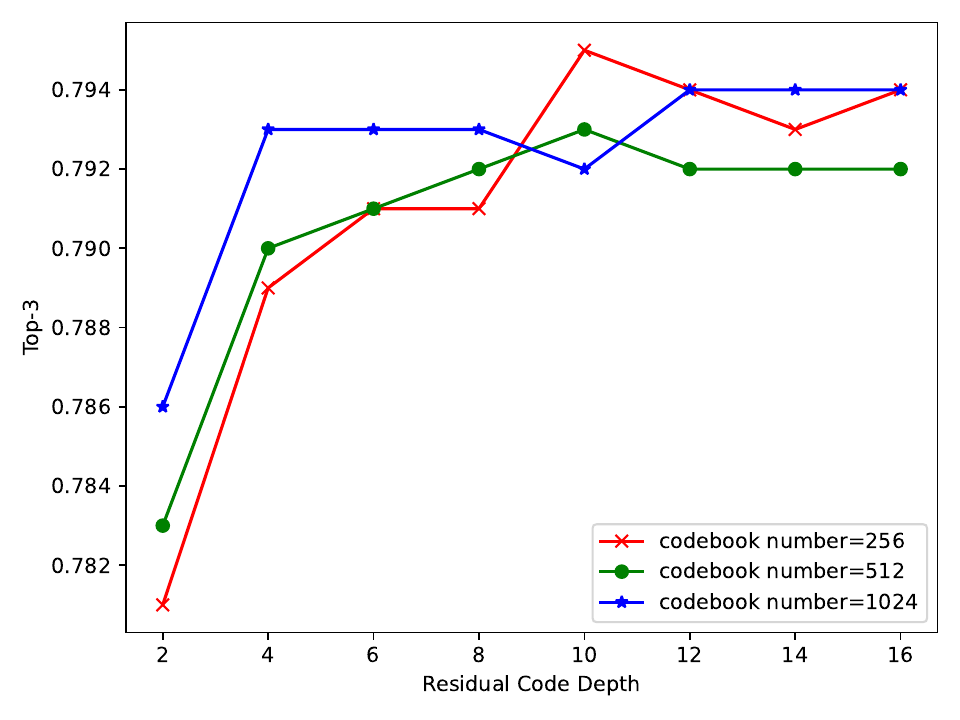}
     \caption{\textbf{Top-3 accuracy}}
 \end{subfigure}
 
\caption{Evaluation on different combinations of residual code depths and codebook sizes.}
\vspace{-10pt}
\label{fig:code_depth_and_codebook_size}
\end{figure}

\paragraph{Evaluation on the choice of residual code depth and codebook size.} We  have investigated the impact of different residual code depths and codebook sizes on the motion reconstruction quality. The results are displayed in ~\cref{fig:code_depth_and_codebook_size}. From the figure, we can see that with a codebook size as small as 256 and large residual code depth such as 10 or 12, the reconstuction quality has already exceeded previous VQ-VAE methods. Simply increasing the codebook size or the residual code depth has margin effects or even leads to overfitting on the limited training dataset. In our experiments, we tend to choose a small codebook size and a large residual code depth for better generalization performance because the small codebook size indicates small class numbers in the generation stage and large code depth indicates increasing the available training code indices due to the usage of \textbf{shared} codebook. Accordingly, for a limited dataset, this combination can compress the representation space, alleviate the overfitting problem and boost the generalization ability of our generation model.

\paragraph{Evaluation on the importance of the corrupted RVQ augmentation.} We then investigate the impact of corrupted RVQ strategies presented in Section 3.4. The results are illustrated in~\cref{tab:ab_corrupt} at the generation stage. We notice that per-time corruption achieves the best results when compared with no-corruption or per-code corruption. The corrupted noise is important in reducing exposure bias in autoregressive models. This can force the model to learn robust latent features. Moreover, for our RVQ-VAE model, the corruption must be in temporal granularity, i.e., the double-tier GPT must "see" some ground truth code indices per time step; otherwise, the performance can still drop. Therefore, the per-time RVQ code corruption strategy has a positive effect in the training phase.

\begin{table}
\centering
\caption{Evaluation on different corrupted RVQ strategies.}\vspace{5pt}
\begin{tabular}{cccc}
\toprule
 Corruption Strategy & FID$\downarrow$ & MM-Dist$\downarrow$ & Top-1$\uparrow$ \\
\midrule
w/o corruption & 0.258 & 3.846 & 0.374\\
\midrule
w per-code corruption & 0.149 & 3.133 & 0.476\\
\midrule
w per-time corruption & \textbf{0.134} & \textbf{3.077} & \textbf{0.486}\\
\bottomrule
\end{tabular}
\label{tab:ab_corrupt}
\end{table}

\begin{figure}[t]
 \centering
 \begin{subfigure}[b]{0.475\columnwidth}
     \centering
     \includegraphics[width=\textwidth]{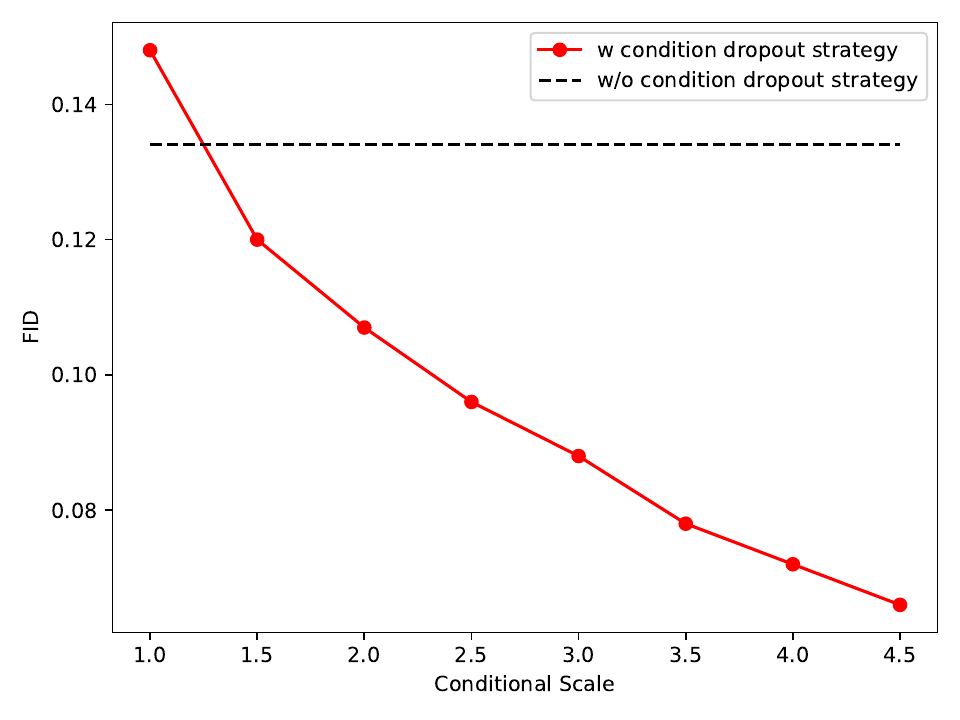}
     \caption{\textbf{FID}}
 \end{subfigure}
 \hfill
 \begin{subfigure}[b]{0.475\columnwidth}
     \centering
     \includegraphics[width=\textwidth]{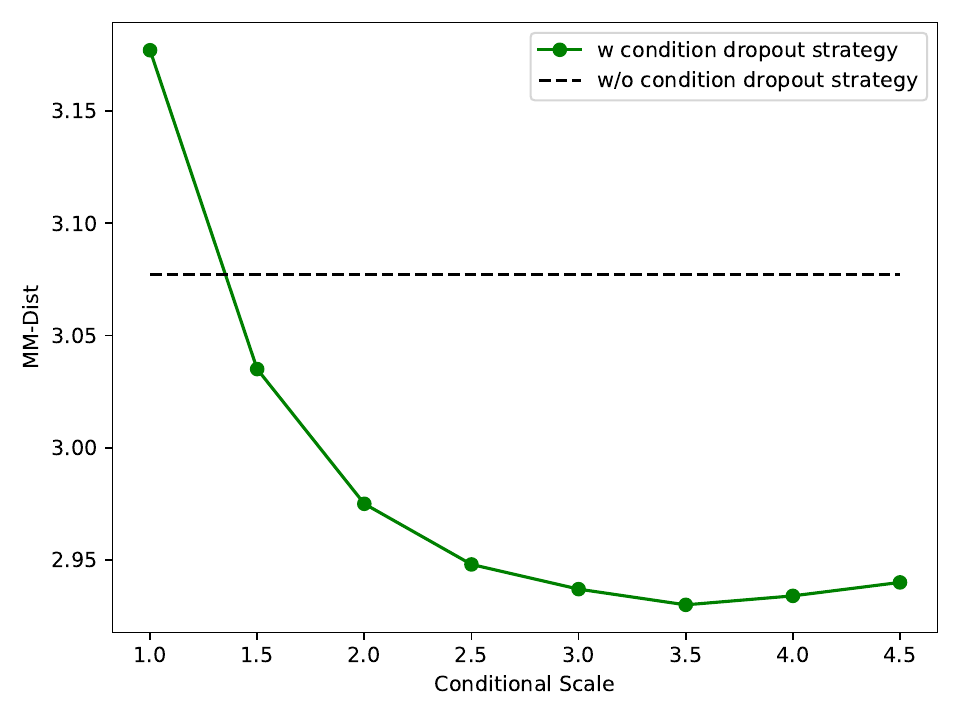}
     \caption{\textbf{MM-Dist}}
 \end{subfigure}
 \hfill
 \begin{subfigure}[b]{0.475\columnwidth}
     \centering
     \includegraphics[width=\textwidth]{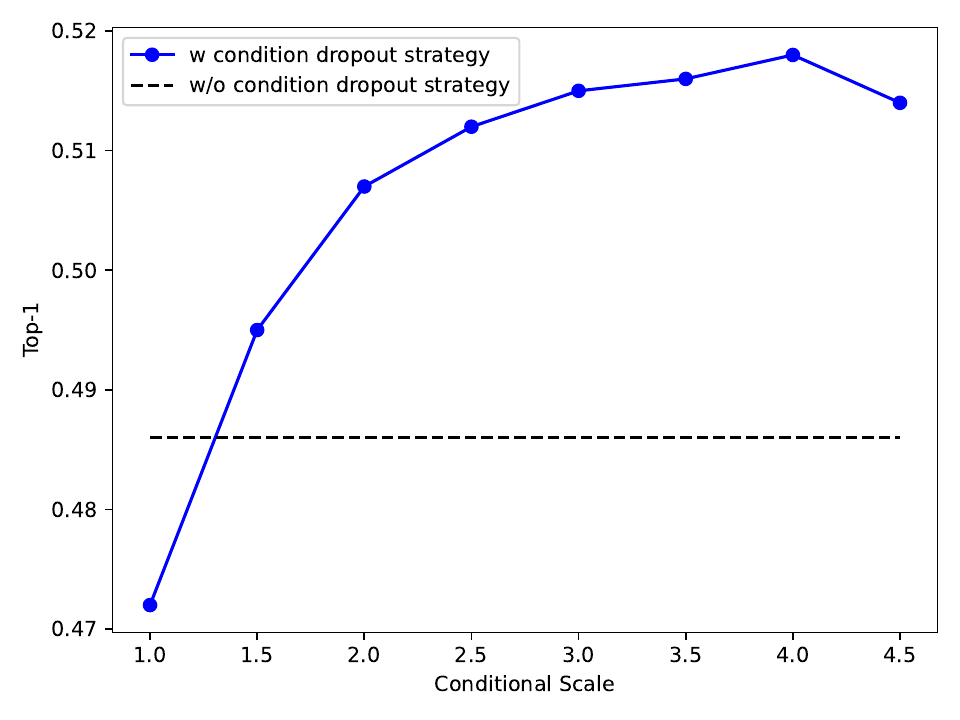}
     \caption{\textbf{Top-1 accuracy}}
 \end{subfigure}
 \hfill
 \begin{subfigure}[b]{0.475\columnwidth}
     \centering
     \includegraphics[width=\textwidth]{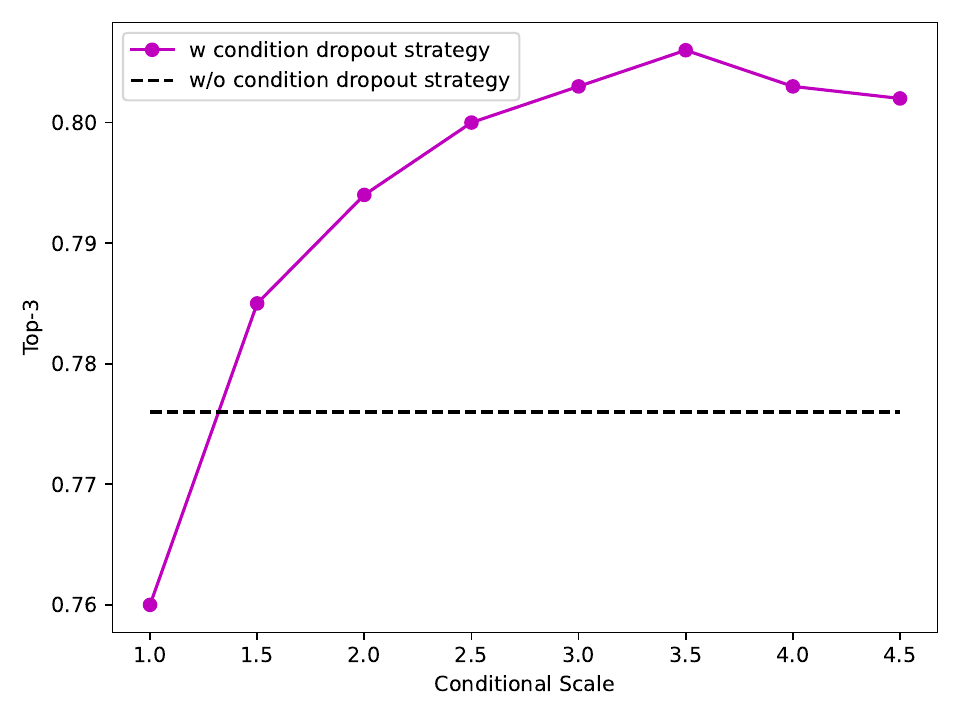}
     \caption{\textbf{Top-3 accuracy}}
 \end{subfigure}
 
\caption{Evaluation on the impact of different conditional scale parameters. The dotted line indicates the results without the conditional dropout and classifier-free guidance strategy.}
\vspace{-10pt}
\label{fig:ab_cfg_tau}
\end{figure}

\paragraph{Evaluation on the condition dropout and classifier-free guidance strategy.}
During the training phase, we randomly drop the CLIP condition vectors and learn the unconditional GPT concurrently within the same model. This strategy is often adopted in the diffusion model to improve conditional matching accuracy. During the inference phase, the classifier-free guidance technique is adopted to compute the mixed logits for probability sampling. In our experiments, we leave out the condition drop strategy and obtain the corresponding results in~\cref{fig:ab_cfg_tau}. From the figure, we can see that the condition dropout strategy and classifier-free guidance can indeed improve the evaluation metric by a large margin.~\cref{fig:ab_cfg_tau} also reports the effects of different conditional scale parameters. As expected, with the sacrifice of sampling diversity, a large conditional scale value such as 4 or 4.5 can greatly boost the synthesis quality, while the text-to-motion matching accuracy seems to be saturated (note that it is already close to the real motion data upon the matching metrics). However, this is a common trade-off in most generative models.

\begin{figure}[t]
 \centering
 \begin{subfigure}[b]{0.475\columnwidth}
     \centering
     \includegraphics[width=\textwidth]{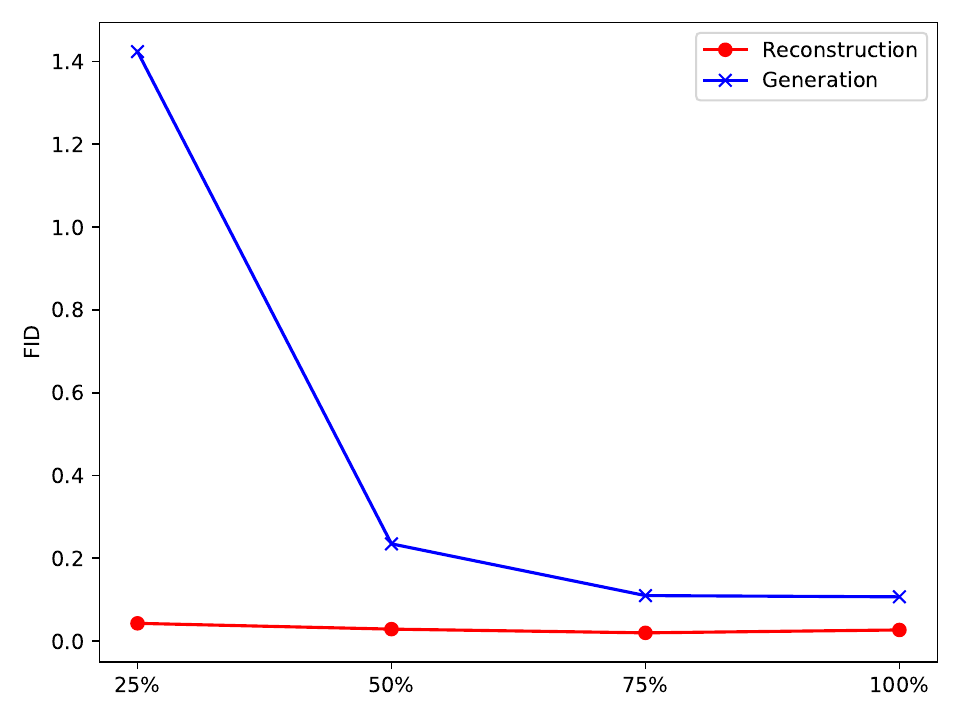}
     \caption{\textbf{FID}}
 \end{subfigure}
 \hfill
 \begin{subfigure}[b]{0.475\columnwidth}
     \centering
     \includegraphics[width=\textwidth]{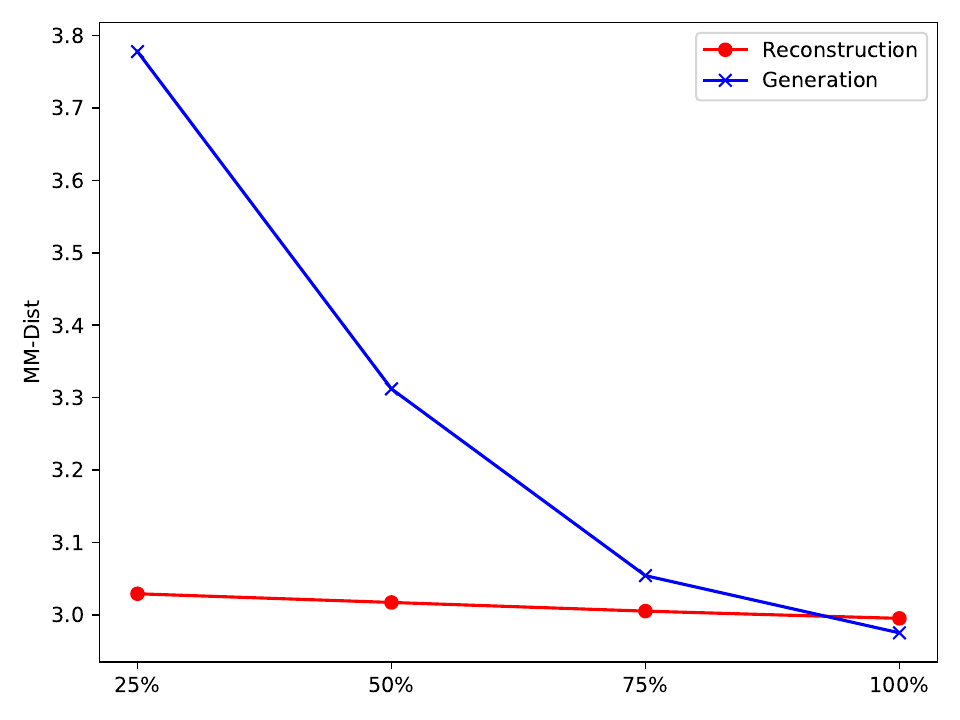}
     \caption{\textbf{MM-Dist}}
 \end{subfigure}
 \hfill
 \begin{subfigure}[b]{0.475\columnwidth}
     \centering
     \includegraphics[width=\textwidth]{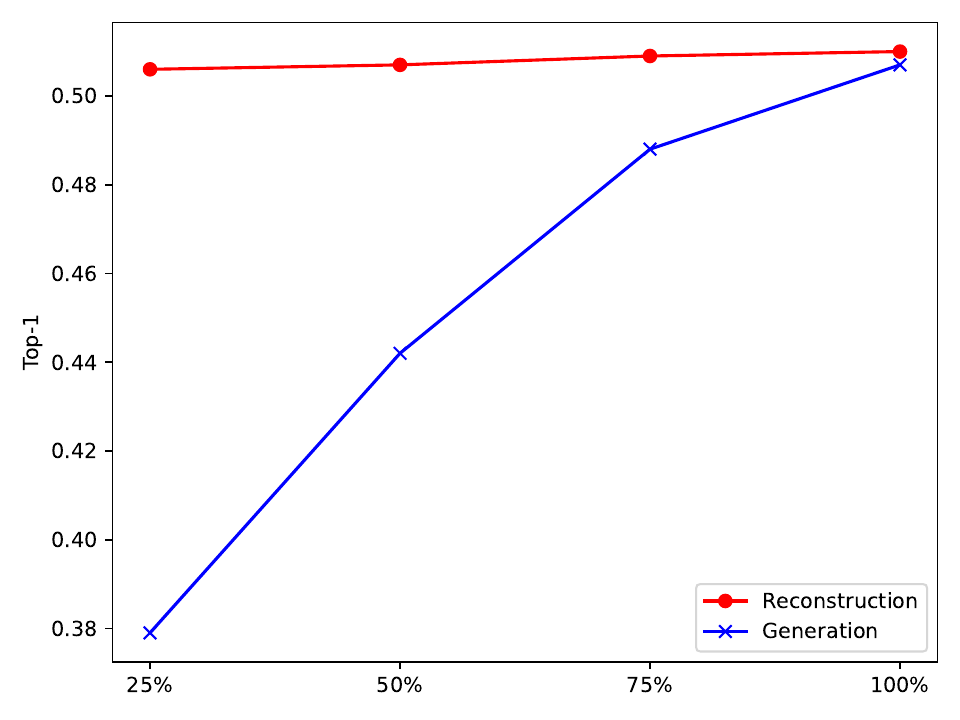}
     \caption{\textbf{Top-1 accuracy}}
 \end{subfigure}
 \hfill
 \begin{subfigure}[b]{0.475\columnwidth}
     \centering
     \includegraphics[width=\textwidth]{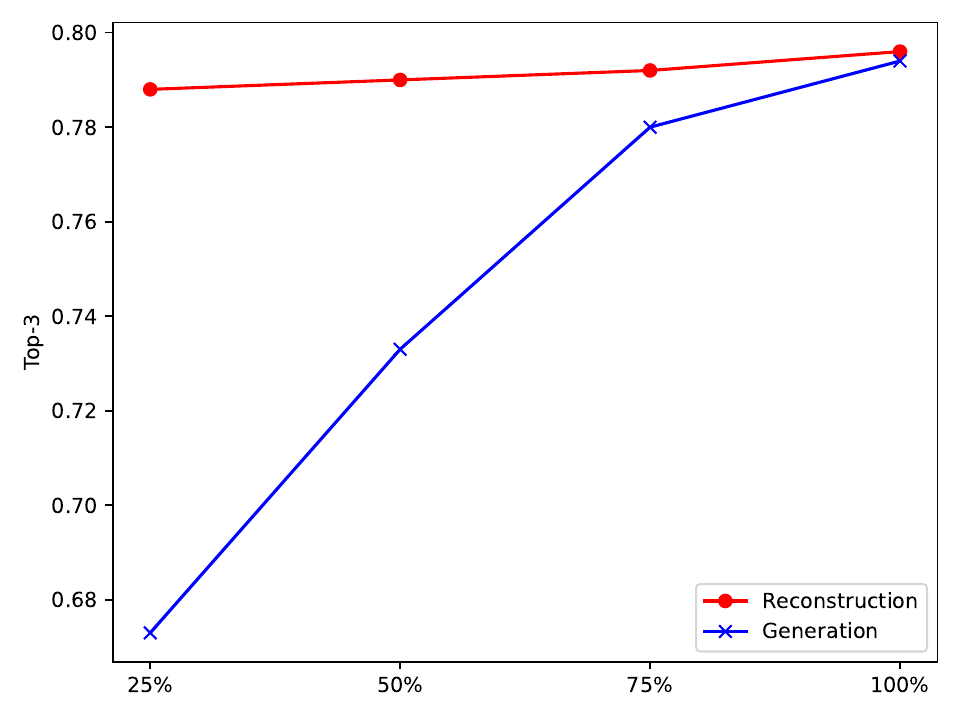}
     \caption{\textbf{Top-3 accuracy}}
 \end{subfigure}
 
\caption{Evaluation on the impact of different dataset sizes.}
\label{fig:ab_dataset}
\end{figure}

\paragraph{Evaluation on the impact of dataset size.} To demonstrate the scalable of our T2M-HiFiGPT, we train the double-tier GPT model with 25\%,50\%,75\%,100\% training data set and observe the corresponding reconstruction and generation performance on the full test set. The scaling results are shown in~\cref{fig:ab_dataset}. From the figure, we can observe the increasing performance when the dataset size grows, which demonstrates the potential performance gain from a larger dataset size. However, the reconstruction stage seems to be saturated when the dataset size grows to 75\%, which indicates the data efficiency and strong generalization ability of the learned RVQ-VAE model. The generation stage shows similar patterns except for the text-to-motion matching accuracy. Overall, this experiment indicates that our method is scalable for large dataset sizes.

\section{Conclusion}
\label{sec:conclusion}
In this work, we propose a novel framework based on RVQ-VAE and double-tier GPT to represent and synthesize human motion data from textual descriptions. Our method achieved SOTA performance when compared against concurrent GPT-based and diffusion-based approaches, suggesting that the proposed framework remains a very competitive approach for the motion generation task. We conducted a large number of ablation investigations and conducted detailed analysis of the two-stage training strategies on human motion reconstruction and generation. Our work provides a novel perspective for the task of high-fidelity human motion generation. Moreover, our framework is very general and may be adopted in other related applications such as facial animation or cloth animation generation, which remains our future work.

{\small
\bibliographystyle{ieee_fullname}
\bibliography{motion}
}

\end{document}